\documentclass[sigconf,9pt]{acmart}
\AtBeginDocument{%
  }

\usepackage{acmart-taps}
\usepackage{stfloats}
\usepackage{multirow}
\usepackage{enumitem}

\usepackage{colortbl}
\definecolor{bestgreen}{RGB}{220,242,220}
\definecolor{secondyellow}{RGB}{255,243,205}
\definecolor{headergray}{RGB}{238,240,244}
\definecolor{methodgray}{RGB}{245,247,250}
\definecolor{sectiongray}{RGB}{245,246,248}
\definecolor{totalgray}{RGB}{235,237,240}

\makeatletter

\def\month@english{\ifcase\month\or
    Jan\or Feb\or Mar\or Apr\or May\or Jun\or
    Jul\or Aug\or Sep\or Oct\or Nov\or Dec\fi}
\makeatother

\newcommand{\method}{\emph{PhysioLite}}

\setcopyright{cc}
\setcctype{by}
\copyrightyear{2026}
\acmYear{2026}
\acmDOI{10.1145/3774906.3802787}
\acmISBN{979-8-4007-2309-4/26/05}
\acmConference[SenSys '26]{ACM/IEEE International Conference on Embedded Artificial Intelligence and Sensing Systems}{May 11--14, 2026}{Saint Malo, France}
\acmBooktitle{ACM/IEEE International Conference on Embedded Artificial Intelligence and Sensing Systems (SenSys '26), May 11--14, 2026, Saint Malo, France}

\begin{document}

\title[Towards Real-Time ECG and EMG Modeling on \textmu NPUs]{Towards Real-Time ECG and EMG Modeling on \ensuremath{\mu}NPUs}

\author{Josh Millar}
\orcid{0009-0002-2247-1594}
\affiliation{\institution{Imperial College London}
\city{London}
\country{United Kingdom}}
\email{jm4622@ic.ac.uk}

\author{Ashok Samraj Thangarajan}
\orcid{0000-0001-9999-7928}
\affiliation{\institution{Nokia Bell Labs}
\city{London}
\country{United Kingdom}}
\email{ashoksamraj.t86@gmail.com}

\author{Soumyajit Chatterjee}
\orcid{0000-0001-5604-2267}
\affiliation{\institution{Brave Software \& University of Cambridge}
\city{London}
\country{United Kingdom}}
\email{sjituit@gmail.com}

\author{Hamed Haddadi}
\orcid{0000-0002-5895-8903}
\affiliation{\institution{Imperial College London}
\city{London}
\country{United Kingdom}}
\email{h.haddadi@imperial.ac.uk}

\renewcommand{\shortauthors}{Millar et al.}

\begin{abstract}
% The miniaturisation of neural processing units (NPUs) and other low-power accelerators has enabled their integration into microcontroller-scale wearable hardware, supporting near real-time, offline, and privacy-preserving inference on-device. However, this has not yet been effectively translated to physiological signal processing, where recent Transformer-based models achieve state-of-the-art performance but are too large for resource and power-constrained hardware and incompatible with \textmu NPUs due to their dynamic attention operations. We present PhysioLite, a lightweight, NPU-compatible convolutional architecture for physiological signal analysis. Using learnable wavelet filter banks, CPU-offloaded positional encoding, and hardware-aware layer design, PhysioLite achieves performance comparable to Transformer baselines while being approximately 10× smaller (<512 KB with 8-bit quantization) on benchmark ECG and EMG datasets. We additionally profile component-wise performance and resource consumption, offering a thorough evaluation to support practical adoption. By reducing memory and compute overheads, PhysioLite enables efficient near real-time inference on the MAX78000 \textmu NPU, demonstrating the feasibility of advanced physiological signal processing on wearable devices. 
The miniaturisation of neural processing units (NPUs) and other low-power accelerators has enabled their integration into microcontroller-scale wearable hardware, supporting near-real-time, offline, and privacy-preserving inference. Yet physiological signal analysis has remained infeasible on such hardware; recent Transformer-based models show state-of-the-art performance but are prohibitively large for resource- and power-constrained hardware and incompatible with \textmu NPUs due to their dynamic attention operations. We introduce \method, a lightweight, NPU-compatible model architecture and training framework for ECG/EMG signal analysis. Using learnable wavelet filter banks, CPU-offloaded positional encoding, and hardware-aware layer design, \method{} reaches performance comparable to state-of-the-art Transformer-based foundation models on ECG and EMG benchmarks, while being <10\% of the size ($\sim$370KB with 8-bit quantization). We also profile its component-wise latency and resource consumption on both the MAX78000  and HX6538 WE2 \textmu NPUs, demonstrating its viability for signal analysis on constrained, battery-powered hardware. We release our model(s) and training framework at: \url{https://github.com/j0shmillar/physiolite}.
\end{abstract}




%\copyrightyear{2026}
%\acmYear{2026}
%\setcopyright{cc}
%\setcctype{by}
%\acmConference[SenSys '26]{ACM/IEEE International Conference on Embedded Artificial Intelligence and Sensing Systems}{May 11--14, 2026}{Saint Malo, France}
%\acmBooktitle{ACM/IEEE International Conference on Embedded Artificial Intelligence and Sensing Systems (SenSys '26), May 11--14, 2026, Saint Malo, France}
%\acmDOI{10.1145/3774906.3802787}
%\acmISBN{979-8-4007-2309-4/2026/05}

\maketitle

%\raggedbottom

\section{Introduction}\enlargethispage{4pt} 
Physiological time-series signals, such as electrocardiography (ECG) and electromyography (EMG), are rich markers of cardiovascular and neuromuscular function, and underpin a wide range of applications, including continuous health monitoring~\cite{intro1}, early disease detection~\cite{intro2}, prosthetic control~\cite{intro3}, and human–computer interaction~\cite{intro4}. The growing interest in personalized and ubiquitous healthcare has motivated the development of wearable systems capable of directly modeling physiological signals, enabling low-latency feedback, user privacy guarantees, and robust deployment in settings with limited or intermittent connectivity. However, physiological signals acquired in real-world environments are often non-stationary, subject-specific~\cite{subject_invariant}, and susceptible to noise and motion artifacts~\cite{artefacts}; effective physiological models must extract and integrate subtle, multi-scale temporal dependencies. Designing such models while satisfying the compute and memory constraints of wearable ($e.g.$, microcontroller-class) hardware remains arduous.

Recent foundation models for physiological signal modeling have adopted Transformer-based architectures, achieving state-of-the-art performance on ECG and EMG benchmarks by capturing long-range temporal dependencies and multi-scale structure~\cite{physiowave, ecgfounder, ecgfm}. Importantly, foundation-scale models exhibit strong generalization across downstream applications; large-scale pretraining enables the learning of broadly applicable features, which can then be adapted to specific clinical or application-driven tasks via fine-tuning. 

Other recent work has investigated specialized embedded platforms for biosignal acquisition and processing. BioGAP and BioGAP-Ultra, for example, are modular, NPU-enabled wearable systems designed for multimodal physiological signal processing and inference across signals including ECG, EMG, and EEG~\cite{frey2023biogap, frey2025biogapultra}. 
Additionally, open-source RISC-V–based platforms have been studied for medical-grade EMG processing, showing that dedicated custom instruction extensions and tightly integrated accelerators can substantially improve efficiency for embedded biosignal processing~\cite{kartsch2025riscv}. Such works demonstrate the growing viability of embedded hardware for biosignal workloads, and motivate the use of hardware-efficient model architectures.

However, such models typically operate on the scale of tens of millions of parameters and rely on memory- and compute-intensive attention mechanisms, making them impractical to execute on resource-constrained wearables~\cite{mcuformer}. Even microcontroller-class neural processing units (\textmu NPUs), while offering hardware \nobreak acceleration, provide only limited on-chip memory and support for a constrained set of static computation kernels. They do not directly support dynamic attention or large intermediate activation buffers. Consequently, state-of-the-art physiological foundation models rely on server- or mobile-grade hardware, limiting their applicability for continuous, private, and real-time inference on wearable devices. 

% \begin{figure}[htbp!]
%   \centering
%   \includegraphics[width=0.95\columnwidth]{figs/highlight2.pdf}
%   \caption{Left - \method{} yields (slightly) improved mean F\textsubscript{1} score with over $10\times$  fewer parameters than PhysioWave. Right - the MAX78000 MCU with onboard \textmu NPU (size compared to a €1 coin).}\label{fig:key}
%   \vspace{-0.2cm}
% \end{figure}
\begin{figure}[t!]
  \centering
  \includegraphics[width=0.7\columnwidth]{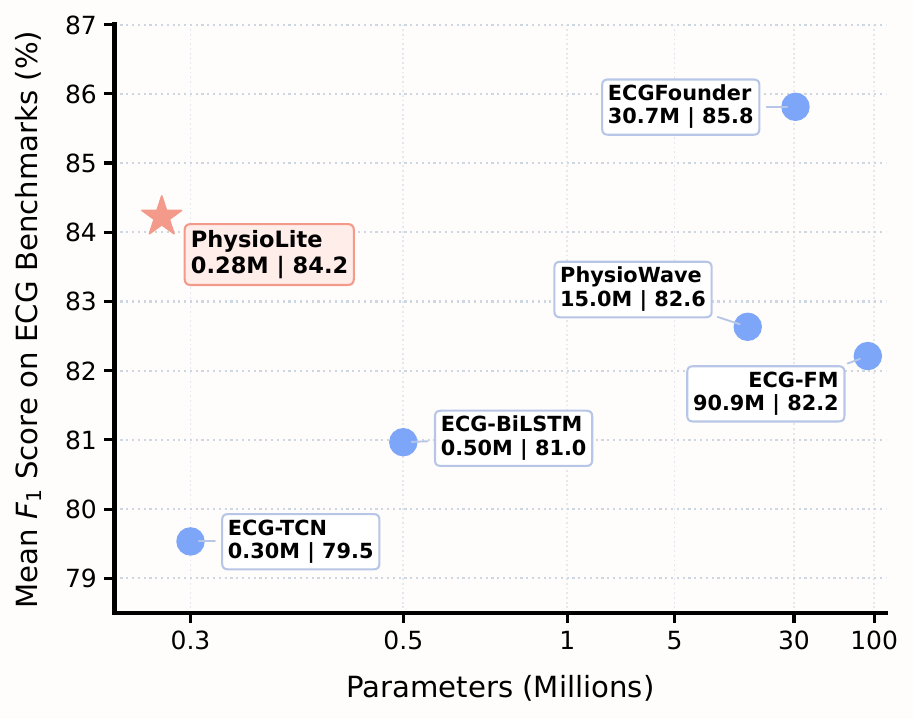}
  \vspace{-0.4cm}
  \caption{Comparison of avg. F\textsubscript{1} on ECG benchmarks (\S\ref{section:results}) with \method{} and a number of ECG foundation models.}\label{fig:key}
  \vspace{-0.5cm}
\end{figure}
We introduce \method, a lightweight architecture and training paradigm designed to bridge the gap between large ECG/EMG foundation models and microcontroller-class hardware. The key idea is to retain the multi-scale temporal modeling capacity and contextual awareness of wavelet-Transformer models~\cite{physiowave}, while replacing hardware-incompatible, or accelerator-unfriendly, attention mechanisms with structured convolutional alternatives and lightweight positional encoding. Specifically, \method{} approximates learnable wavelet decomposition using parallel convolution branches to capture frequency-specific dynamics, and substitutes attention blocks with compact feature-mixing modules that rely on pointwise convolutions. Positional information is externally incorporated on the CPU prior to accelerator execution, ensuring temporal context is maintained without requiring dynamic operations on the accelerator. This design mirrors the functional behavior of attention-based foundation models while remaining compatible with \textmu NPU operator sets and strict memory constraints, enabling local inference under tight memory and power budgets.

To assess the effectiveness of our design, we evaluate \method{} across a suite of ECG and EMG benchmark datasets commonly used for supervised classification and arrhythmia or gesture recognition tasks. Compared to state-of-the-art ECG and EMG foundation models~\cite{physiowave}, \method{} remains competitive while staying under a 512KB memory budget ($\sim$370KB with 8-bit quantization, <0.3M parameters), as shown in \figurename~\ref{fig:key}. On PTB-XL~\cite{ptb_xl}, CPSC~\cite{cpsc_2018}, and Chapman--Shaoxing~\cite{chapman_shaoxing} ECG benchmarks, \method{} achieves AUROC of 0.893--0.994 and F\textsubscript{1} of 0.600--0.942, nearing foundation-scale models \cite{ecgfounder, ecgfm}. On Ninapro DB5~\cite{ninapro_db5}, EPN-612~\cite{epn612}, and UCI~\cite{uci_emg} sEMG gesture benchmarks, \method{} achieves F\textsubscript{1} scores of 0.738--0.949 and AUROC of 0.994--0.996, matching or exceeding prior foundation models while using substantially fewer parameters, demonstrating that key multi-scale physiological structures can be captured without RoPE attention \cite{rope}, or other attention mechanisms. Finally, through end-to-end deployment and component-wise profiling on the MAX78000~\cite{max78000} and Himax HX6538 WE2~\cite{himax} \textmu NPUs, we show that \method{} fits within on-chip memory limits and achieves near real-time latency (sub $\sim$20~ms), confirming its suitability for ECG and EMG monitoring on such hardware. 

To summarize, our key contributions are:
\begin{enumerate}[topsep=1mm, itemsep=0mm, parsep=0mm, left=0mm] 
\item We detail \method, a lightweight and \textmu NPU-compatible model architecture and training framework for general-purpose ECG and EMG processing, designed to preserve the representational strengths of Transformer-based foundation models, %, while avoiding hardware-incompatible attention mechanisms. 
 and release our pretrained model(s) and framework as open-source:\\ \url{https://github.com/j0shmillar/physiolite}.
\item We conduct a comprehensive evaluation of our method across multiple ECG and EMG benchmarks, showing that \method{} achieves performance comparable to state-of-the-art foundation models while reducing model size and compute footprint by approximately an order of magnitude.
\item We perform detailed system-level profiling on the MAX78000 and Himax WE2, \textmu NPU-integrated microcontrollers designed for battery-powered wearables. We quantify latency, memory usage, and power consumption, across various processing stages, and demonstrate near-real-time inference latency (sub $\sim$20~ms) suitable for continuous monitoring.
\end{enumerate}

\section{Related Work and Background}

\subsection{\textmu NPU Hardware Constraints} 

NPUs are specialized hardware accelerators designed for executing neural networks. In the last few years, low-power microcontroller-scale variants, \textmu NPUs, have emerged, supporting neural acceleration on battery-powered and wearable hardware. 

\begin{figure}[htbp!]
   \vspace{-0.3cm}
  \centering
  \includegraphics[width=0.9\columnwidth]{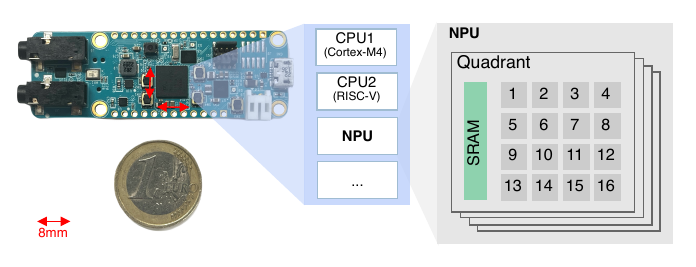}
  \vspace{-0.4cm}
  \caption{The MAX78000-FTHR alongside a €1 coin for scale. The on-chip NPU architecture consists of a systolic array organized into four quadrants, each containing 16 processing elements (64 total), with dedicated per-quadrant SRAM. 
  }\label{fig:npu_arch}
  \vspace{-0.3cm}
\end{figure}

Fig. \ref{fig:npu_arch}, below, outlines the general design and layout of a \textmu NPU, composed of a systolic array of processing elements (PEs)~\cite{npu_arch}. Each PE holds its own multiply-accumulate units and, importantly, its own dedicated weight SRAM; this avoids memory contention during network inference and exploits the inherent parallelism of CNN layers and dense matrix multiplications. The array of PEs is linked with a global buffer and DRAM via an on-chip communication grid. \textmu NPUs have been shown to yield order-of-magnitude speedups in inference latency compared with CPU-only execution on equivalent-scale hardware~\cite{benchmark}. 

% The general layout of a \textmu NPU is composed of a systolic array of processing elements (PEs)~\cite{npu_arch}. Each PE holds its own multiply-accumulate units and, importantly, its own dedicated weight SRAM; this avoids memory contention during network inference and exploits the inherent parallelism of CNN layers and dense matrix multiplications. The array of PEs is linked with a global buffer and DRAM via an on-chip %communication 
% grid. \textmu NPUs have been shown to yield order-of-magnitude speedups in inference latency compared with CPU-only execution on equivalent-scale hardware~\cite{benchmark}. 

\textmu NPUs impose strict hardware constraints. They generally provide only limited on-chip memory space for model storage and I/O ($e.g.$, hundreds of KB), forcing heavy model quantization and other hardware-specific optimizations. The MAX78000’s accelerator, for example, only includes $\sim$442 KB of weight SRAM (with support for 1-8 bit model precision)~\cite{max78000}. The majority of platforms also currently have hugely limited operator support, heavily optimized for convolutional layers and a small set of accompanying functions, such as pooling. They often impose additional constraints on layer parameterizations, such as kernel configurations, and enforce specific operator configurations for fusion. NPUs are also generally optimized for static weights and layers, lacking support for layers with dynamic behaviours ($e.g.$, self-attention). Vendors are addressing this constraint; for example, Arm has recently built support for general matrix multiply operations into its Ethos-U85 NPU~\cite{arm_ethos_u85}. However, dynamic weight support is unlikely to appear in microcontroller-scale NPUs for some time.

In summary, current \textmu NPUs are \emph{optimized for static weights} and layers, \emph{lacking support for dynamic behaviours such as self-attention}.

\begin{figure}[htbp!]
  \vspace{-0.3cm}
  \centering
  \includegraphics[width=0.8\columnwidth]{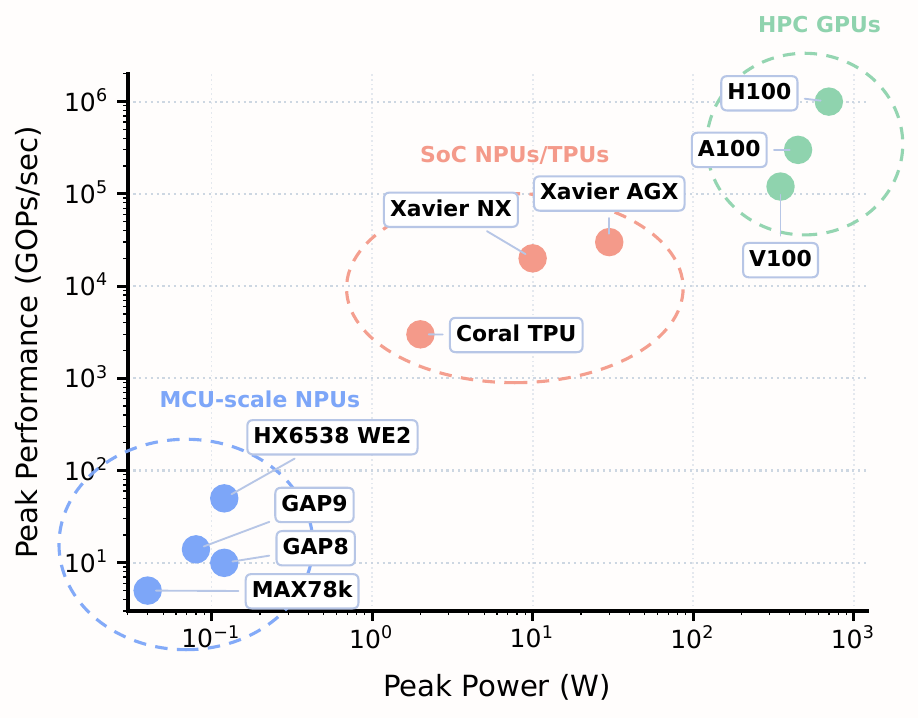}
  \vspace{-0.4cm}
  \caption{Performance of representative hardware accelerators. Microcontroller (MCU)-scale NPUs are uniquely suited for low-power, wearable systems \cite{npu_comp}.
  }\label{fig:comp_plot}
  \vspace{-0.5cm}
\end{figure}

\subsection{Hardware-aware Model Design}
Executing neural models on constrained hardware offers well-recognized advantages, including predictable latency, improved data privacy through offline operation, and reduced deployment costs for model vendors. Accordingly, a substantial body of work has explored optimizing inference efficiency through lightweight architecture design---including with neural architecture search (NAS)---and dynamic or mixed-precision quantization~\cite{opt1, opt2, opt3}. However, most of these efforts target CNN- or RNN-based architectures~\cite{ml_4_mcu_review}. Transformer models remain considerably more memory-intensive: their self-attention layers involve large all-to-all matrix multiplications and nonlinear operations that are inherently incompatible with typical embedded accelerators. Several recent works have attempted to adapt Transformers to microcontroller-class hardware. MCUFormer~\cite{mcuformer}, for instance, employs one-shot NAS, patch-wise streaming, and low-rank decomposition to fit a ViT, achieving over 70\% ImageNet accuracy, within 320~KB of SRAM on an STM32F7. Jung \textit{et~al.} leverage parallelization and loop reordering on a GAP9 cluster~\cite{gap9,jung}, while Dequino \textit{et~al.} introduce a runtime library exploiting attention-weight redundancy for efficient deployment of lightweight ViTs~\cite{dequino}.

In brief, while \emph{most early work on hardware-aware optimizations for microcontrollers focused on CNNs and RNNs, recent efforts have also extended to Transformer models.}

\subsection{ECG and EMG Signal Modeling} 
Recently, PhysioWave~\cite{physiowave}, a multi-scale wavelet-driven architecture for general physiological signal analysis, achieved state-of-the-art results on multiple downstream physiological applications. Its architecture follows a compact ViT-inspired design: a learnable wavelet decomposition captures localized %frequency 
content across multiple temporal scales, attention blocks model long-range temporal dependencies, and rotary positional embeddings %explicitly 
encode temporal relationships.  

Hybrid models with CNN front-ends and Transformer encoders exhibit similar performance across specific downstream applications, albeit with a larger model size. DeepECG-Net, for instance, showcases real-time inference on a Cortex-A72 CPU %~\cite{cortexA72} 
 with <50ms latency and $\sim$30MB memory usage~\cite{deepecgnet}. WaveFormer~\cite{waveformer}, a streamlined Transformer-based model for sEMG-based applications, also targets low-power CPUs. It captures time- and frequency-domain content using a novel, learnable %convolution-based 
front-end that applies multi-level decompositions via depthwise-separable convolutions. Using only key EMG frequency patterns and parameter-efficient attention blocks, WaveFormer demonstrates >90\% accuracy on the %612-subject 
EPN-612 benchmark, with only $\sim$3.1million parameters. It runs at sub-7ms latency on an i7-11800H with 8-bit quantization. 

% In this work, we build on both PhysioWave’s wavelet-based feature extraction and the hardware-aware design approaches from WaveFormer, towards \emph{supporting general-purpose physiological signal analysis on \textit{microcontroller-scale} (\textmu NPU) hardware}.

\section{Designing \method}
% In this section, we outline the overall design of \method{}, and how it improves over PhysioWave to become compatible for \textmu NPUs. We begin by highlighting the key limitations of PhysioWave and explaining why it is incompatible with deployment on \textmu NPUs. Then we proceed with the development of different components of \method{} in a hardware-aware manner. 

\begin{figure}[htbp!]
  \vspace{-0.2cm}
  \centering
  \includegraphics[width=1.0\columnwidth]{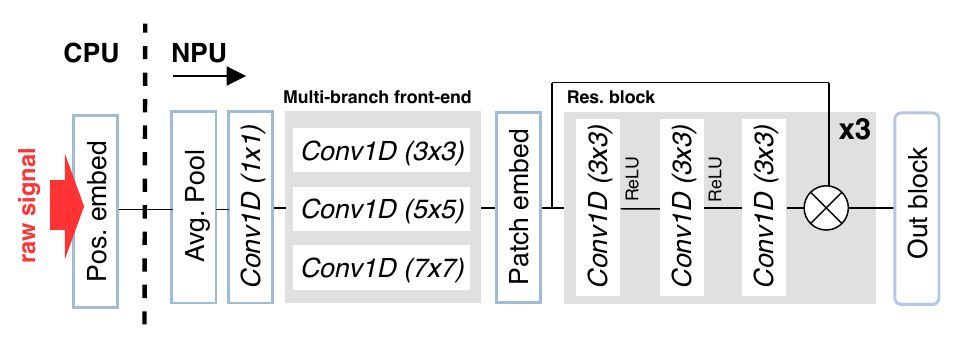}
  \vspace{-0.9cm}
  \caption{\method{}'s architecture mapped to CPU and NPU execution. Multi-branch temporal convolutions provide \textit{wavelet-like} feature extraction.
  }\label{fig:system}
  \vspace{-0.3cm}
\end{figure}

Deploying ECG/EMG models on constrained hardware requires much redesign; attention, adaptive wavelet selection, and learnable positional encoding are all either unsupported or computationally infeasible on microcontroller-class NPUs. The attention operations in particular rely on large matrix multiplications with dynamic, data-dependent weights, while \textmu NPU instruction sets are designed for convolutions with static kernels. \method{} retains the general architectural structure of physiological foundation models---with a multi-scale front-end, explicit temporal encoding, and subsequent feature mixing---while replacing hardware-incompatible components with convolutional, statically defined, NPU-compliant operators. Figure \ref{fig:system} outlines \method{}'s hybrid CPU–NPU architecture.

\subsection{Source Foundation Model}
% \textbf{PhysioWave:  An Overview}
We adopt a distillation-based training paradigm and use PhysioWave as our source model. PhysioWave is a recent foundation model for general-purpose ECG and sEMG modeling. Its architecture comprises three main components: (i) a learnable wavelet decomposition front-end that factorizes raw signals into multi-scale, frequency-aware sub-bands; (ii) explicit positional encoding to preserve temporal structure; and (iii) multi-head self-attention (MHSA) blocks that aggregate information across time and scales. The decomposition front-end employs low- and high-pass filters initialized from discrete wavelets and refined via data-driven optimization, producing scale-separated sub-bands that are processed in parallel to capture features at multiple temporal resolutions. This design is well-suited to physiological signals, exhibiting characteristic multi-scale patterns, such as P–QRS–T complexes in ECG or short muscle-activation bursts in sEMG, while also requiring modeling of longer-range temporal dependencies that single-scale convolutions fail to capture. We leverage PhysioWave as our source model to transfer its learned wavelet filters, enabling the \nobreak processing of raw physiological signals end-to-end, avoiding prohibitively expensive bandpass filtering and preprocessing on the hardware.

% \noindent\textbf{\textcolor{red}{Limitations of ECG/EMG Foundation Models:}} Deploying ... on constrained hardware requires much redesign; attention, adaptive wavelet selection, and learnable positional encoding are all either unsupported or computationally infeasible on microcontroller-class NPUs. The attention operations in particular rely on large matrix multiplications with dynamic, data-dependent weights, while \textmu NPU instruction sets are designed for convolutions with static kernels. \method{} retains the architectural structure of ...---a multi-scale front-end, explicit temporal encoding, and subsequent feature mixing---while replacing hardware-incompatible components with convolutional, statically defined, NPU-compliant operators. Figure \ref{fig:system} below outlines \method{}'s hybrid CPU–NPU architecture.

\subsection{CPU-Based Positional Encoding}
Continuous learnable embeddings, relying on high-precision floating-point arithmetic, are incompatible with \textmu NPU instruction sets. We instead offload positional encoding to the CPU: positional information is generated as deterministic sinusoidal functions on the host processor, then concatenated with the raw input before being moved to the NPU. To make positional embedding generation efficient, we precompute sine and cosine lookup tables (LUTs) for a small set of reusable base periods, eliminating redundant evaluations. The LUTs are quantized to the Q7 format, allowing the CPU to stream embeddings directly into the NPU input buffers with minimal arithmetic overhead. This introduces a minor latency overhead---profiling indicates embedding generation accounts for $\sim$11\% of end-to-end latency---but remains essential, as we find performance drops sharply when positional encodings are removed. The embedding at position $t$ and frequency index $k$ is defined as:
\vspace{0.15cm}
\[
PE_{k}(t) = 
\begin{bmatrix}
\sin\left(2\pi \cdot 2^{k} \cdot \frac{t}{T}\right), &
\cos\left(2\pi \cdot 2^{k} \cdot \frac{t}{T}\right)
\end{bmatrix},
\quad k = 0, \ldots, F-1
\]

Here, $t$ denotes the index within a window of length $T$, $k$ the frequency index, and $F$ the number of sinusoidal frequency pairs used. In practice, we scale the positional channels by a small factor (e.g., $\alpha \sim 0.1$) to stabilize early-layer activations and prevent the positional signal from dominating the core physiological features.

% \subsection{Learnable Wavelet-like Filter Banks}

% PhysioWave’s front-end applies a learnable wavelet decomposition: low- and high-pass filters are initialized from discrete wavelets, adapted via data-driven optimization, and applied over multiple levels to produce scale-separated sub-bands. Each sub-band is then processed by parallel branches, capturing features at multiple temporal resolutions and frequency scales. The filter coefficients are progressively refined via an unsupervised signal reconstruction process.

% We emulate this decomposition using a hardware-friendly approximation based on parallel convolution branches. Each branch applies a different kernel size (3, 5, 7), corresponding to distinct receptive fields, capturing temporal structure at multiple resolutions. Prior to this, a lightweight stem convolution performs initial channel mixing. This configuration allows \method{} to approximate wavelet-like decompositions using only standard convolution operators, fully compatible with embedded accelerators. The resulting representation retains the core behaviour of PhysioWave’s multi-scale filtering, disentangling frequency-specific dynamics while remaining computationally efficient.

% The front-end directly operates on the raw (windowed) multi-channel signal; this is ideal for constrained hardware, as it avoids costly pre-processing or signal-filtering stages such as bandpass, notch, or baseline-wander removal.

\subsection{Learnable Wavelet-like Filter Banks}

\method{} approximates wavelet decomposition using a hardware-efficient multi-branch convolutional front-end. A lightweight stem convolution first mixes channels, followed by parallel branches with different kernel sizes (3, 5, and 7) to capture temporal structure at multiple resolutions. This design mimics the multi-scale filtering of wavelet transforms: each branch implements a learnable finite impulse response filter whose receptive field corresponds to a specific subband, allowing the model to approximate low-, mid-, and high-frequency components without explicit wavelet bases~\cite{wavelet_conv3, wavelet_conv2, wavelet_conv_1}. Operating directly on the raw, windowed multi-channel signal, the front-end removes the need for hand-crafted preprocessing such as bandpass filtering or baseline removal, enabling deployment on resource-constrained \textmu NPU hardware.

% \subsection{Emulating MHSA with Feature Mixing}

% The most substantial departure from PhysioWave occurs in its feature mixing layers. PhysioWave’s strength derives from its attention blocks, which enable global feature aggregation across multiple temporal scales and resolutions. However, as outlined above, attention operations rely on large matrix multiplications with dynamic, data-dependent weights, which cannot be effectively compiled into \textmu NPU instruction sets designed for convolutions with static kernels.
% To address this, our architecture replaces MHSA with a series of pointwise convolutional bottlenecks, approximating the feature mixing and non-linear behaviour of PhysioWave’s attention and MLP sub-blocks. Each bottleneck applies an inverted residual structure to capture fine-grained temporal dependencies within its receptive field, while our positional encodings convey the global contextual information full attention would otherwise learn implicitly. This design reflects our key insight: for 1-D signal processing, local convolutional mixing combined with explicit, multi-scale positional encodings can emulate the representational effects of global attention.

\subsection{Feature Mixing without Attention}
The final component of our design replaces self-attention with an efficient convolutional feature-mixing mechanism. Instead of MHSA, \method{} employs a sequence of pointwise convolutional bottlenecks that act as lightweight token mixers. This approach is supported by prior work demonstrating that depthwise–pointwise separable bottlenecks are powerful channel mixers despite their low computational cost. MobileNet’s inverted residual blocks~\cite{mobilenetv2} show that such bottlenecks provide strong representational capacity even in highly resource-constrained settings. More broadly, a growing body of ``attention-free” architectures has shown that the core ViT scaffold remains effective when MHSA is replaced with simpler mixing operations, including MLPs and convolutional layers. MetaFormer~\cite{metaformer}, for example, demonstrates that token mixing can be achieved with pointwise layers followed by non-linear transformations. Other works show that depthwise–pointwise convolutions approximate the behavior of MHSA~\cite{patches_are_all_you_need, convnet2020s}.

In \method{}, the bottleneck layers adopt an inverted residual structure to capture fine-grained temporal dependencies within their receptive fields, while multi-scale positional encodings provide the global context that attention would otherwise model dynamically. This combination approximates the non-linear mixing behavior associated with MHSA using only convolutional operators that are fully compatible with \textmu NPU instruction sets.

\section{Experimental Setup}
% \sumo{Do we have a picture of the testbed? IMO it really helps if we can give a picture of the testbed or even the hardware in the paper. I have seen reviewers comparing the size and other features.}
We benchmark \method{} across multiple datasets on both ECG and EMG modalities, spanning a range of downstream clinical and human–computer interaction applications. 
Table \ref{tab:model-summary} summarizes our benchmark models. We use PhysioWave's 5M configuration, fine-tuned, as our source model for the EMG datasets, and its 15M configuration for ECG. We also implement and evaluate against a custom TCN-style, \textmu NPU-friendly convolutional network (\emph{ECG/EMG-TCN}), for quantifying the gains from our architectural design beyond foundation models, alongside a BiLSTM-style recurrent network (\emph{ECG/EMG-BiLSTM}).

% \sumo{I will rewrite this as follows. We profile resource consumption on two ultra-low-power \textmu NPU platforms. First, we use the MAX78000 \cite{max78000}, which integrates a Cortex-M4F and a RISC-V core with a proprietary 30-GOPS CNN accelerator. This accelerator provides dedicated on-chip SRAM—512~KB for input activations, 442~KB for weights, and 2~KB for biases—and supports quantized inference at 1-, 2-, 4-, and 8-bit precision. The MAX78000 also includes 512~KB of flash and 128~KB of CPU-accessible SRAM.
% Second, we evaluate on the Himax HX6538 WE2 (HX-WE2) \cite{himax}, a Corstone-300–based \textmu NPU platform combining a Cortex-M55 CPU with an Ethos-U55 NPU, offering up to 512~GOPS. It provides 512~KB TCM, 2~MB SRAM, and 16~MB flash.}
We profile resource consumption on the MAX78000 \cite{max78000}, a low-power \textmu NPU featuring both a Cortex-M4F and a RISC-V processor, alongside a proprietary 30-GOPS CNN accelerator. The latter has a dedicated 512 KB SRAM for input data, 442 KB for weights, and 2 KB for biases, and supports quantized operations at 1, 2, 4, and 8-bit precision. The MAX78000 also has 512 KB of flash and 128 KB of CPU-only SRAM. We also evaluate on the Himax HX6538 WE2 (or HX-WE2) \cite{himax}, a Corstone-300 \textmu NPU set up, with Cortex-M55 CPU and Ethos-U55 NPU, delivering up to 512 GOPS. The platform features 512KB TCM, 2MB SRAM, and 16MB flash. We additionally profile end-to-end latency using a CPU-only execution pipeline on the HX-WE2 to quantify the performance gains attributable specifically to NPU acceleration.

\subsection{Datasets}

\begin{table*}[t]
\centering
\caption{Summary of datasets used in our ECG and EMG benchmarks. \\ \#S = number of subjects/patients; Hz = sampling frequency; Ch = number of channels.}
\vspace{-0.3cm}
\label{tab:dataset-summary}
\centering
\scriptsize
\begin{tabular}{llllccl}
\toprule
\textbf{Type} & \textbf{Dataset} & \textbf{Task} & \textbf{\#S} & \textbf{Hz} & \textbf{Ch} & \textbf{Loc.} \\
\midrule
\multirow{3}{*}{ECG}
& PTB-XL~\cite{ptb_xl} & Multi-label cardiac classification & 18885 & 500 & 12 & 12-lead clinical ECG \\
& Chapman--Shaoxing~\cite{chapman_shaoxing} & Arrhythmia / rhythm classification & 45152 & 500 & 12 & 12-lead clinical ECG \\
& CPSC 2018~\cite{cpsc_2018} & Arrhythmia detection & 10330 & 500 & 12 & 12-lead clinical ECG \\
\midrule
\multirow{3}{*}{EMG}
& Ninapro DB5~\cite{ninapro_db5} & Gesture recognition (52 moves + rest) & 10 & 200 & 16 & Forearm, 2$\times$ Myo~\cite{myo} armbands \\
& EPN-612~\cite{epn612} & Gesture recognition (6 moves) & 612 & 200 & 8 & Forearm, Myo armband \\
& UCI EMG~\cite{uci_emg} & Gesture recognition (10 moves) & 36 & 200 & 8 & Forearm\\
\bottomrule
\end{tabular}
\vspace{-0.3cm}
\end{table*}

\begin{table}[t]
\centering
\caption{Summary of models used in our ECG and EMG benchmarks. N Params = number of trainable parameters. \\ SL = supervised, SSL = self-supervised.}
\vspace{-0.3cm}
\label{tab:model-summary}
\scriptsize
\resizebox{\columnwidth}{!}{
\begin{tabular}{lllll}
\toprule
\textbf{Type} & \textbf{Model} & \textbf{N Params} & \textbf{Training} & \textbf{Pretrain} \\
\midrule

\multirow{2}{*}{ECG}
& ECGFounder & 30.7M & SL (multi-label, 150 diagnoses) & HEEDB (10{,}771{,}552 ECGs) \cite{heedb} \\
& ECG-FM & 90.9M & SSL (w/ hybrid objective) & 1.4M ECG (\cite{physionet}, \cite{mimic}, 0.6M priv.) \\
\midrule

\multirow{2}{*}{EMG}
& WaveFormer & 3.1M & SL & N/A \\
& OTiS & 45M & SSL & Multi-domain time series corpus \\
\midrule

\multirow{3}{*}{Both}
& PhysioWave & 5M / 15M / 37M & SSL (masked recon) & Large ECG + EMG corpus \\
& ECG/EMG-TCN & 0.3M & SL & N/A \\
& ECG/EMG-BiLSTM & 0.5M & SL & N/A \\
& \method{} & <0.3M & SL & N/A \\
\bottomrule
\end{tabular}
}
\vspace{-0.5cm}
\end{table}

Table \ref{tab:dataset-summary} summarizes our benchmark ECG and EMG datasets. For single-label datasets (DB5, EPN612, UCI-EMG, PTB-XL) we report macro F\textsubscript{1}. For multilabel datasets (CPSC, Chapman-Shaoxing) we report micro F\textsubscript{1} on thresholded predictions.

\subsubsection{Data Pre-Processing}
ECG signals are split into overlapping windows of 2048 samples, and EMG into 1024 samples with a step size of 512. For Ninapro DB5, we use a 512-sample window with a step size of 64; we retain transition periods between DB5 gestures to emulate real-time modeling, reducing label noise. ECG signals are resampled to 500~Hz, while EMG signals are sampled at 200~Hz. PTB ECG windows are normalized using per-window MinMax scaling, while other ECG and EMG signals are z-score normalized. Recordings with insufficient lead counts are zero-padded.

% \newpage

For models that require it, we apply explicit signal conditioning prior to normalization. For ECG, each lead is (i) notch filtered at 50~Hz (IIR notch, $Q{=}30$, zero-phase), (ii) band-pass filtered using a 4th-order Butterworth filter from 0.67--40~Hz (zero-phase), and (iii) baseline-wander removed by subtracting a median-filter estimate of the baseline (kernel length 0.4~s). For EMG, each channel is band-pass filtered using a 4th-order Butterworth filter from 20~Hz to $0.95\times$Nyquist (zero-phase; i.e., up to 950~Hz at 2~kHz), followed by a 50~Hz IIR notch filter ($Q{=}30$, zero-phase) to suppress powerline interference. Both signals are then z-score normalized.

We use subject-wise data partitioning for each dataset. % We also utilize random dropout, jitter, and amplitude scaling on DB5. 

\subsubsection{On the Hardware.} 
Preprocessing on the hardware includes five stages: (1) resampling, (2) zero-padding, (3) per-channel z-score normalization, (4) quantization to Q7, and (5) concatenation with learned positional embeddings. Each step is optimized for our limited on-chip SRAM and integer-oriented architectures.
Resampling aligns input signals acquired at arbitrary ADC sampling rates with the model’s expected temporal resolution (e.g., 2 kHz). This is achieved via an in-place linear interpolator and %single 
per-channel scratch buffer, avoiding memory duplication across channels. Padding is applied to ensure each window has fixed length. %(i.e., $T$ = 1024 or 2048).

Per-channel z-score normalization is computed in floating-point using a streaming accumulation of the mean and variance, followed by normalization and scaling into an 8-bit Q7 format. The normalized and quantized signals are then combined with precomputed sinusoidal positional embeddings. %and passed to the NPU. 
This is done using the CPU with memory-efficient buffer reuse to avoid SRAM overflow.

Unlike conventional approaches to EMG/ECG processing, ours omits explicit, computationally intensive signal-filtering steps, such as bandpass, notch, or baseline-wander removal \cite{filtering1, filtering2}. These operations are instead learned implicitly by the model’s first stage, a set of learnable wavelet-like filter banks. %realized as depthwise convolutions. %The resulting design supports continuous, low-power monitoring under strict memory constraints.

\subsection{Hardware Measurement Environment}

We record latency, power, and memory usage for 1024- and 2048-sample EMG and ECG applications. Models are compiled and deployed for the MAX78000 using its \texttt{ai8x} \cite{ai8x} compiler. For the HX-WE2, ARM's Vela \cite{vela} compiler is used. Vela applies Ethos-U-specific optimizations and supports both \textit{Size} and \textit{Performance} optimization strategies. We use the \textit{Performance} optimization strategy to optimize inference latency using the available arena cache. We use the TFLM runtime for CPU-only profiling on the HX-WE2 \cite{tflm}.

We measure latency, broken down into the various preprocessing and inference stages, using each platform’s internal clock, and repeat the measurements across 10 runs. We use each NPUs default runtime configuration. With the MAX78000, only the Cortex-M4 CPU and the CNN accelerator (NPU) are active. Both run at their maximum operating frequencies (100 MHz and 50 MHz, respectively). We enforce the same CPU/NPU frequencies on the HX-WE2. 

We measure power using a high-voltage power monitor \cite{monsoon} at a sampling rate of 5 kHz, and also repeat measurements over 10 runs. The input voltage is fixed at 3.3 V. Warmup period is set at 60s. %The memory usage is drawn from the linker (\textit{.map}) file generated by each compiler.

\subsection{Hyperparameter \& Distillation Setup}

We distill \method{} from the downstream PhysioWave model fine-tuned on each application (ECG and EMG). We used a model depth of 3, embedding dimension of 256, and pre-concatenated 1D sinusoidal positional channels (8/12 frequencies) to encode temporal structure. Our distillation approach combines a standard loss with a distillation loss using temperature-scaled logits:
\vspace{-0.1cm}
\[
\mathcal{L} = (1 - \alpha_{\text{kd}})\,\mathcal{L}_{\text{hard}} + \alpha_{\text{kd}}\,\mathcal{L}_{\text{KD}},
\]

$e.g.$, with \(\alpha_{\text{kd}} = 0.3\) and \(T = 2.0\).

Optimization used AdamW~\cite{adamw} (\(lr = 1\times10^{-3}\), weight decay \(= 1\times10^{-3}\)) with a cosine learning-rate schedule and linear warmup over 5 epochs. We used 30 epochs total, and a batch size of 16. Our hard loss used binary cross-entropy with logits for multi-label ECG and standard cross-entropy for EMG. Hyperparameter configurations for our \textit{baseline} models are detailed in Appendix \ref{sec:hyperparameters}.
We use cross-entropy for single-label datasets, binary cross-entropy for multi-label, and cross-entropy with a soft macro-F1 (one-vs-rest) regularization term on the imbalanced DB5 and PTB-XL datasets.

\vspace{-0.1cm}

\section{Results}
\subsection{Performance on Downstream Applications}
\label{section:results}

\noindent\textbf{ECG.}
\tablename~\ref{tab:ecg-results} summarizes how \method{} compares against both foundation-scale and compact sequence models across our ECG benchmarks. On Chapman–Shaoxing, \method{} exhibits strong performance with 0.942~F\textsubscript{1} and 0.994~AUROC, improving upon ECG-BiLSTM (0.929~F\textsubscript{1}) and nearing ECG-FM (0.955~F\textsubscript{1}) and ECG-TCN (0.959~F\textsubscript{1}), but remains measurably below ECGFounder (0.970~F\textsubscript{1}, 0.997~AUROC). On PTB-XL, \method{} reaches 0.600~F\textsubscript{1} and 0.893~AUROC, exceeding PhysioWave and compact sequence models such as ECG-TCN (0.527~F\textsubscript{1}) and ECG-BiLSTM (0.577~F\textsubscript{1}), but again remains below the large pretrained foundation models such as ECGFounder (0.645~F\textsubscript{1}, 0.919~AUROC). This is likely due to their substantially greater representational capacity coupled with large-scale pretraining. Notably, ECG-TCN also slightly exceeds \method{} on Chapman–Shaoxing despite similar parameter counts; this dataset is large (45k+ patients) and rhythm-centric, and we hypothesize that dilated temporal convolutions can capture its rhythm structure and long-range dependencies without multi-scale decomposition. \method{}’s wavelet-inspired multi-branch front-end and µNPU-constrained design also introduces structural regularization, and may marginally limit performance on such large, homogeneous datasets. For CPSC, \method{} reaches 0.772~F\textsubscript{1} and 0.962~AUROC, improving upon ECG-FM (0.706~F\textsubscript{1}) and ECGFounder in F\textsubscript{1} (0.758). Across the ECG benchmarks, \method{} improves upon most compact pretrained models, while remaining competitive with the large-scale foundation models.

We also quantify the impact of key architectural components %, with ablations provided in Appendix \ref{sec:ablations}. 
(see Appendix \ref{sec:ablations}). Across ECG benchmarks, the use of distillation improves mean F\textsubscript{1} by +3.8\% (0.786$\rightarrow$0.825). The impact of removing positional encoding is dataset-dependent, with PTB-XL and Chapman--Shaoxing showing a measurable F\textsubscript{1} drop, while CPSC improves from 0.719 to 0.766~F\textsubscript{1} (+5.6\%). Our multi-scale kernel design yields improved mean F\textsubscript{1} compared to fixed kernels: +0.3\% vs $K=3$, +0.8\% vs $K=5$, and +0.5\% vs $K=7$ (0.767 vs 0.764/0.759/0.762), with PTB-XL and Chapman--Shaoxing as exceptions. %(0.605, $K=3$ vs 0.600, $K={3,5,7}$).

% \vspace{1em}
\noindent\textbf{EMG.}
\tablename~\ref{tab:emg-results} summarizes how \method{} compares against state-of-the-art sEMG, and both large-scale foundation and compact sequence models, across our EMG benchmarks.

On Ninapro DB5, \method{} reaches 0.738~F\textsubscript{1} and 0.994~AUROC, surpassing PhysioWave (0.736~F\textsubscript{1}, 0.993~AUROC). WaveFormer’s lightweight attention blocks provide modest improvements in terms of F\textsubscript{1} score (0.761~F\textsubscript{1}), likely due to its ability to focus on fine-grained transition boundaries across the 52 gesture classes. For EPN-612, \method{} reaches 0.949~F\textsubscript{1} and 0.996~AUROC, almost matching our compact sequence models, EMG-TCN (0.951~F\textsubscript{1}) and EMG-BiLSTM (0.954~F\textsubscript{1}), and improving upon WaveFormer (0.928~F\textsubscript{1}) and OTiS (0.913~F\textsubscript{1}). On UCI-EMG, \method{} achieves 0.939~F\textsubscript{1} and 0.995~AUROC, comparable to EMG-BiLSTM (0.938~F\textsubscript{1}) and OTiS (0.939~F\textsubscript{1}), while outperforming PhysioWave (0.926~F\textsubscript{1}). %Across our EMG benchmarks, \method{} remains competitive with state-of-the-art.  
On both EPN-612 (612 subjects) and UCI-EMG (36 subjects), the compact models (i.e. EMG-TCN and EMG-BiLSTM) match or slightly exceed \method{}; these datasets contain relatively short, low-variance gesture segments, and we hypothesize that simpler sequential inductive biases are sufficient and multi-scale modeling may not confer additional benefit. In such regimes, \method{}’s broader multi-branch receptive-field design and distillation may act as mild over-regularization.

Distillation improves mean F\textsubscript{1} by $\sim$4.6\%, with the largest gains observed on DB5 (+5.6\%) and UCI-EMG (+7.3\%). %, suggesting supervision is particularly beneficial for datasets with fine-grained activation patterns. 
Removing positional encodings produces a mean drop of approximately 6.6\% in F\textsubscript{1}, with the largest drops on DB5 ($-$13.1\%) and UCI-EMG ($-$10.4\%). In contrast, EPN-612 shows only a minor decrease ($-$0.3\%), indicating reduced dependence on temporal ordering. This reflects structural differences across datasets: EPN-612 is dominated by shorter, lower-variance signal segments. Curiously, similar to CPSC, removing positional encodings slightly improves performance relative to the non-KD model, suggesting that explicit positional bias may be unnecessary when discriminative features are largely invariant.% to temporal alignment. 

\aptLtoX[graphic=no,type=html]{\begin{table}[t]
    \centering
    \setlength{\tabcolsep}{3.5pt}
    \renewcommand{\arraystretch}{1.12}
    \scriptsize
        \caption{\method{} vs baselines across our ECG benchmarks. Best results are highlighted in green and second-best in yellow.}
        \vspace{-0.25cm}
        \label{tab:ecg-results}
        \centering
        \resizebox{\textwidth}{!}{
        \begin{tabular}{l ccc ccc ccc}
            \toprule
            \textbf{Model} 
            & \multicolumn{3}{c}{\textbf{PTB-XL}} 
            & \multicolumn{3}{c}{\textbf{CPSC}} 
            & \multicolumn{3}{c}{\textbf{Chapman-Shaoxing}} \\
            \cmidrule(lr){2-4} \cmidrule(lr){5-7} \cmidrule(lr){8-10}
             & \textbf{Acc.} & \textbf{F\textsubscript{1}} & \textbf{AUROC} 
             & \textbf{Acc.} & \textbf{F\textsubscript{1}} & \textbf{AUROC} 
             & \textbf{Acc.} & \textbf{F\textsubscript{1}} & \textbf{AUROC} \\
            \midrule

            PhysioWave 
                & 0.729 & 0.599 & 0.891
                & 0.694 & 0.701 & 0.944
                & 0.947 & 0.946 & \cellcolor{secondyellow}\textbf{0.996} \\
            ECGFounder
                & \cellcolor{bestgreen}\textbf{0.760} & \cellcolor{bestgreen}\textbf{0.649} & \cellcolor{bestgreen}\textbf{0.919}
                & \cellcolor{secondyellow}\textbf{0.736} & \cellcolor{secondyellow}\textbf{0.758} & \cellcolor{bestgreen}\textbf{0.964}
                & \cellcolor{bestgreen}\textbf{0.970} & \cellcolor{bestgreen}\textbf{0.970} & \cellcolor{bestgreen}\textbf{0.997} \\
            ECG-FM
                & 0.689 & 0.560 & 0.870
                & 0.715 & 0.706 & 0.947
                & 0.961 & 0.955 & \cellcolor{secondyellow}\textbf{0.996} \\
            ECG-TCN
                & 0.672 & 0.527 & 0.859
                & 0.610 & 0.639 & 0.939
                & \cellcolor{secondyellow}\textbf{0.964} & \cellcolor{secondyellow}\textbf{0.959} & 0.989 \\
            ECG-BiLSTM
                & 0.712 & 0.577 & 0.875
                & 0.659 & 0.685 & 0.930
                & 0.929 & 0.929 & 0.991 \\

            \method{} 
                & \cellcolor{secondyellow}\textbf{0.731} & \cellcolor{secondyellow}\textbf{0.600} & \cellcolor{secondyellow}\textbf{0.893}
                & \cellcolor{bestgreen}\textbf{0.737} & \cellcolor{bestgreen}\textbf{0.772} & \cellcolor{secondyellow}\textbf{0.962}
                & 0.941 & 0.942 & 0.994 \\
            \bottomrule
        \end{tabular}
        }
\end{table}
\begin{table}
        \caption{\method{} vs baselines across our EMG benchmarks. Best results are highlighted in green and second-best in yellow.}
        \vspace{-0.25cm}
        \label{tab:emg-results}
        \centering
        \resizebox{\textwidth}{!}{
        \begin{tabular}{l ccc ccc ccc}
            \toprule
            \textbf{Model} 
            & \multicolumn{3}{c}{\textbf{DB5}} 
            & \multicolumn{3}{c}{\textbf{EPN612}} 
            & \multicolumn{3}{c}{\textbf{UCI-EMG}} \\
            \cmidrule(lr){2-4} \cmidrule(lr){5-7} \cmidrule(lr){8-10}
             & \textbf{Acc.} & \textbf{F\textsubscript{1}} & \textbf{AUROC} 
             & \textbf{Acc.} & \textbf{F\textsubscript{1}} & \textbf{AUROC} 
             & \textbf{Acc.} & \textbf{F\textsubscript{1}} & \textbf{AUROC} \\
            \midrule

            PhysioWave 
                & \cellcolor{secondyellow}\textbf{0.896} & 0.736 & \cellcolor{secondyellow}\textbf{0.993}
                & 0.940 & 0.940 & 0.995
                & \cellcolor{secondyellow}\textbf{0.929} & 0.926 & \cellcolor{secondyellow}\textbf{0.998} \\

            WaveFormer
                & 0.885 & \cellcolor{bestgreen}\textbf{0.761} & \cellcolor{bestgreen}\textbf{0.994}
                & 0.928 & 0.928 & 0.993         
                & 0.927 & 0.927 & \cellcolor{secondyellow}\textbf{0.998} \\
            OTiS
                & 0.892 & 0.717 & 0.992
                & 0.912 & 0.913 & 0.992
                & 0.905 & \cellcolor{bestgreen}\textbf{0.939} & 0.993 \\
            EMG-TCN
                & 0.876 & 0.685 & 0.992
                & \cellcolor{secondyellow}\textbf{0.951} & \cellcolor{secondyellow}\textbf{0.951} & \cellcolor{bestgreen}\textbf{0.997}
                & 0.917 & 0.915 & \cellcolor{bestgreen}\textbf{0.999} \\
            EMG-BiLSTM
                & 0.804 & 0.489 & 0.982
                & \cellcolor{bestgreen}\textbf{0.954} & \cellcolor{bestgreen}\textbf{0.954} & \cellcolor{bestgreen}\textbf{0.997}
                & \cellcolor{bestgreen}\textbf{0.941} & \cellcolor{secondyellow}\textbf{0.938} & 0.993 \\
            \method{} 
                & \cellcolor{bestgreen}\textbf{0.900} & \cellcolor{secondyellow}\textbf{0.738} & \cellcolor{bestgreen}\textbf{0.994}
                & 0.949 & 0.949 & \cellcolor{secondyellow}\textbf{0.996}
                & \cellcolor{bestgreen}\textbf{0.941} & \cellcolor{bestgreen}\textbf{0.939} & 0.995 \\
            \bottomrule
        \end{tabular}
        }
\end{table}}{
\begin{table*}[t]
    \centering
    \setlength{\tabcolsep}{3.5pt}
    \renewcommand{\arraystretch}{1.12}
    \scriptsize

    \begin{minipage}[t]{0.49\textwidth}
        \caption{\method{} vs baselines across our ECG benchmarks. Best results are highlighted in green and second-best in yellow.}
        \vspace{-0.25cm}
        \label{tab:ecg-results}
        \centering
        \resizebox{\textwidth}{!}{
        \begin{tabular}{l ccc ccc ccc}
            \toprule
            \textbf{Model} 
            & \multicolumn{3}{c}{\textbf{PTB-XL}} 
            & \multicolumn{3}{c}{\textbf{CPSC}} 
            & \multicolumn{3}{c}{\textbf{Chapman-Shaoxing}} \\
            \cmidrule(lr){2-4} \cmidrule(lr){5-7} \cmidrule(lr){8-10}
             & \textbf{Acc.} & \textbf{F\textsubscript{1}} & \textbf{AUROC} 
             & \textbf{Acc.} & \textbf{F\textsubscript{1}} & \textbf{AUROC} 
             & \textbf{Acc.} & \textbf{F\textsubscript{1}} & \textbf{AUROC} \\
            \midrule

            PhysioWave 
                & 0.729 & 0.599 & 0.891
                & 0.694 & 0.701 & 0.944
                & 0.947 & 0.946 & \cellcolor{secondyellow}\textbf{0.996} \\
                
            ECGFounder
                & \cellcolor{bestgreen}\textbf{0.760} & \cellcolor{bestgreen}\textbf{0.649} & \cellcolor{bestgreen}\textbf{0.919}
                & \cellcolor{secondyellow}\textbf{0.736} & \cellcolor{secondyellow}\textbf{0.758} & \cellcolor{bestgreen}\textbf{0.964}
                & \cellcolor{bestgreen}\textbf{0.970} & \cellcolor{bestgreen}\textbf{0.970} & \cellcolor{bestgreen}\textbf{0.997} \\
            ECG-FM
                & 0.689 & 0.560 & 0.870
                & 0.715 & 0.706 & 0.947
                & 0.961 & 0.955 & \cellcolor{secondyellow}\textbf{0.996} \\
            ECG-TCN
                & 0.672 & 0.527 & 0.859
                & 0.610 & 0.639 & 0.939
                & \cellcolor{secondyellow}\textbf{0.964} & \cellcolor{secondyellow}\textbf{0.959} & 0.989 \\
            ECG-BiLSTM
                & 0.712 & 0.577 & 0.875
                & 0.659 & 0.685 & 0.930
                & 0.929 & 0.929 & 0.991 \\

            \method{} 
                & \cellcolor{secondyellow}\textbf{0.731} & \cellcolor{secondyellow}\textbf{0.600} & \cellcolor{secondyellow}\textbf{0.893}
                & \cellcolor{bestgreen}\textbf{0.737} & \cellcolor{bestgreen}\textbf{0.772} & \cellcolor{secondyellow}\textbf{0.962}
                & 0.941 & 0.942 & 0.994 \\
            \bottomrule
        \end{tabular}
        }
    \end{minipage}
    \hfill
    \begin{minipage}[t]{0.49\textwidth}
        \caption{\method{} vs baselines across our EMG benchmarks. Best results are highlighted in green and second-best in yellow.}
        \vspace{-0.25cm}
        \label{tab:emg-results}
        \centering
        \resizebox{\textwidth}{!}{
        \begin{tabular}{l ccc ccc ccc}
            \toprule
            \textbf{Model} 
            & \multicolumn{3}{c}{\textbf{DB5}} 
            & \multicolumn{3}{c}{\textbf{EPN612}} 
            & \multicolumn{3}{c}{\textbf{UCI-EMG}} \\
            \cmidrule(lr){2-4} \cmidrule(lr){5-7} \cmidrule(lr){8-10}
             & \textbf{Acc.} & \textbf{F\textsubscript{1}} & \textbf{AUROC} 
             & \textbf{Acc.} & \textbf{F\textsubscript{1}} & \textbf{AUROC} 
             & \textbf{Acc.} & \textbf{F\textsubscript{1}} & \textbf{AUROC} \\
            \midrule

            PhysioWave 
                & \cellcolor{secondyellow}\textbf{0.896} & 0.736 & \cellcolor{secondyellow}\textbf{0.993}
                & 0.940 & 0.940 & 0.995
                & \cellcolor{secondyellow}\textbf{0.929} & 0.926 & \cellcolor{secondyellow}\textbf{0.998} \\

            WaveFormer
                & 0.885 & \cellcolor{bestgreen}\textbf{0.761} & \cellcolor{bestgreen}\textbf{0.994}
                & 0.928 & 0.928 & 0.993         
                & 0.927 & 0.927 & \cellcolor{secondyellow}\textbf{0.998} \\
            OTiS
                & 0.892 & 0.717 & 0.992
                & 0.912 & 0.913 & 0.992
                & 0.905 & \cellcolor{bestgreen}\textbf{0.939} & 0.993 \\
            EMG-TCN
                & 0.876 & 0.685 & 0.992
                & \cellcolor{secondyellow}\textbf{0.951} & \cellcolor{secondyellow}\textbf{0.951} & \cellcolor{bestgreen}\textbf{0.997}
                & 0.917 & 0.915 & \cellcolor{bestgreen}\textbf{0.999} \\
            EMG-BiLSTM
                & 0.804 & 0.489 & 0.982
                & \cellcolor{bestgreen}\textbf{0.954} & \cellcolor{bestgreen}\textbf{0.954} & \cellcolor{bestgreen}\textbf{0.997}
                & \cellcolor{bestgreen}\textbf{0.941} & \cellcolor{secondyellow}\textbf{0.938} & 0.993 \\
            \method{} 
                & \cellcolor{bestgreen}\textbf{0.900} & \cellcolor{secondyellow}\textbf{0.738} & \cellcolor{bestgreen}\textbf{0.994}
                & 0.949 & 0.949 & \cellcolor{secondyellow}\textbf{0.996}
                & \cellcolor{bestgreen}\textbf{0.941} & \cellcolor{bestgreen}\textbf{0.939} & 0.995 \\
            \bottomrule
        \end{tabular}
        }
    \end{minipage}
    \vspace{-0.3cm}
\end{table*}}

\begin{figure}[htbp!]
  \vspace{-0.3cm}
  \centering
  \includegraphics[width=1.0\columnwidth]{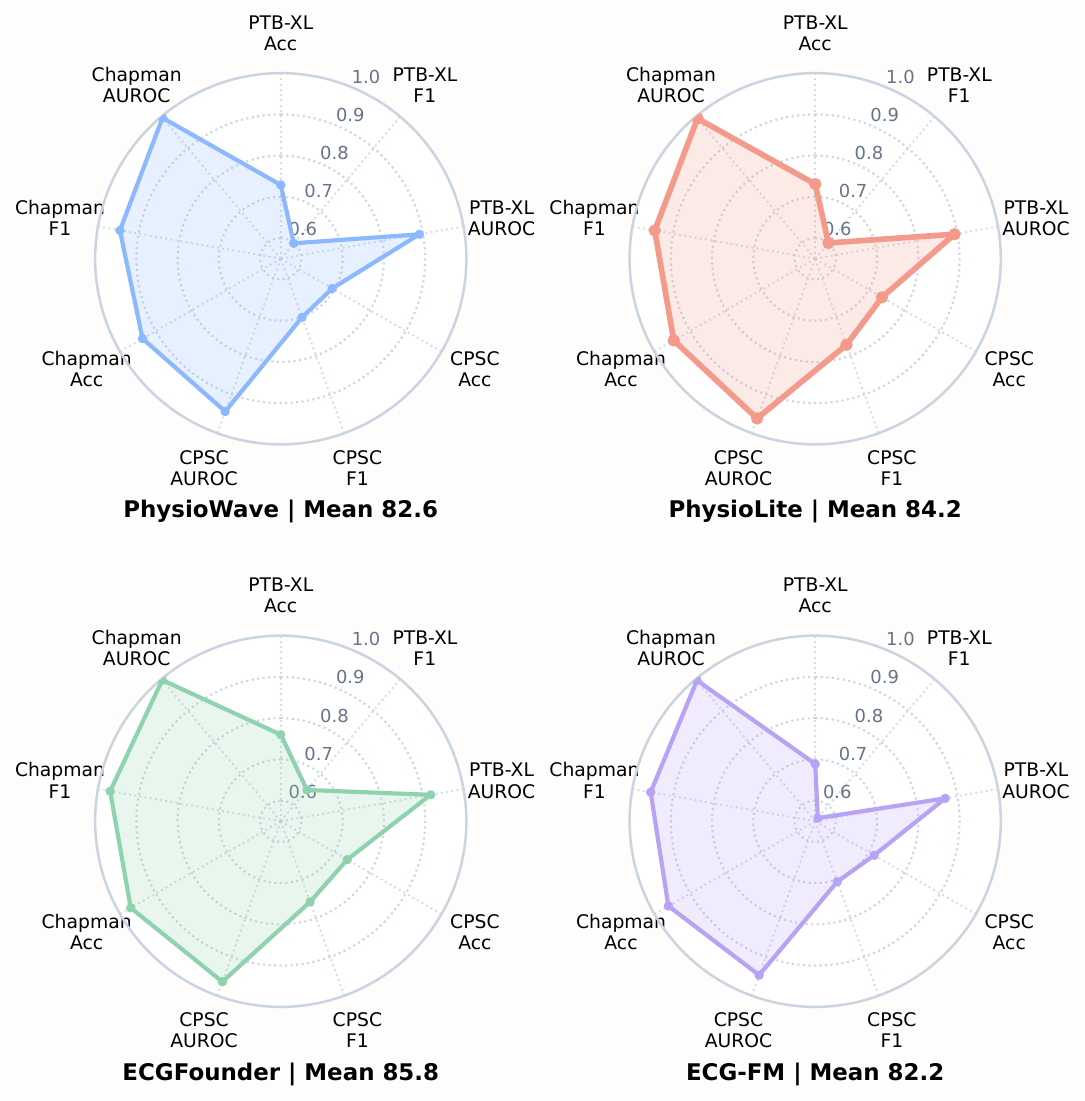}
  \vspace{-0.6cm}
  \caption{Performance comparison across ECG benchmark. Despite its compact size, \method{} yields metrics comparable to foundation-scale models.
  }\label{fig:radar}
  \vspace{-0.4cm}
\end{figure}

\subsection{Hardware Benchmark}
\tablename~\ref{tab:overall} outlines end-to-end latencies of 12.5~ms and 20.3~ms (ECG) and 10.8~ms and 16.9~ms (EMG) for 1024- and 2048-sample size inputs on the MAX78000, and 11.7~ms and 17.5~ms (ECG) and 10.0~ms and 13.9~ms (EMG) on the HX-WE2, respectively. Latency is dominated by z-score normalization and quantization, accounting for roughly 25--50\% of the end-to-end runtime across platforms and window sizes, followed by tile packing and positional encoding. Tile packing and positional encoding contribute an additional 15--20\% depending on window size and platform. CNN inference remains relatively constant across window sizes, taking $\sim$4.6~ms on the MAX78000 and $\sim$4.0~ms on the HX-WE2. CPU-side pre-processing (resampling and z-score/quantization) is the primary bottleneck. CPU-only execution on the HX-WE2 requires 62.3--71.7~ms (ECG) and 64.1--68.0~ms (EMG), showing a $\sim$5--6$\times$ end-to-end speedup from using the NPU, with CNN inference %alone 
accelerated by over 14$\times$.

Memory usage comparison is shown in \tablename~\ref{tab:memory}. The MAX78000 remains within SRAM limits after optimization, with utilization increasing from 43.3~KB (33.1\%) at 1024 samples to 76.1~KB (58.1\%) at 2048 samples. Flash usage on MAX78000 is 370.4~KB (70.7\%) for both window sizes. Flash usage on the HX-WE2 increases from 78.3~KB (29.9\%) at 1024 samples to 123~KB (47.1\%) at 2048 samples, while SRAM utilization remains at 548~KB (27.9\%) for both configurations.

\newpage
Power measurements for the MAX78000 (\tablename~\ref{tab:emg_power}) show a mean consumption of 33.4~mW (peak 35.7~mW) during 2048-sample inference, with 13.4mW when idle, confirming suitability for battery-powered operation. The HX-WE2 exhibits substantially greater power draw, with a mean consumption of 103.8~mW (peak 111.0~mW) during 2048-sample inference and 88.6~mW when idle. Note that, in practice, ECG/EMG inference may be event-driven ($i.e.$, on detected activity), with the NPU largely idle.

\begin{table}[!t]
% \vspace{-0.1cm}
\caption{Memory usage comparison for 1024- and 2048-sample EMG on MAX78000 and HX-WE2.}
\vspace{-0.3cm}
\scriptsize
\centering
\begin{tabular}{cccccc}
\hline
\textbf{Sample} &
\textbf{Platform} &
\textbf{Region} &
\textbf{Region} &
\textbf{Used} &
\textbf{Used} \\
 &  &  & \textbf{Size (KB)} & \textbf{(KB)} & \textbf{(\%)} \\ \hline
\multirow{4}{*}{\centering 1024}
 & \multirow{2}{*}{MAX78000}  & FLASH & 512  & 370 & 70.7 \\
 &                           & SRAM  & 128  & 43.3  & 33.1 \\
 & \multirow{2}{*}{HX-WE2} & FLASH & 256  & 78.3  & 29.9 \\
 &                           & SRAM  & 1924 & 548 & 27.9\\ \hline
\multirow{4}{*}{\centering 2048}
 & \multirow{2}{*}{MAX78000}  & FLASH & 512  & 370 & 70.7 \\
 &                           & SRAM  & 128  & 76.1  & 58.1 \\
 & \multirow{2}{*}{HX-WE2} & FLASH & 256  & 123  & 47.1 \\
 &                           & SRAM  & 1924 & 548 & 27.9 \\ \hline
\end{tabular}
\label{tab:memory}
\vspace{-0.2cm}
\end{table}

\begin{figure}[htbp!]
  \vspace{-0.2cm}
  \centering
  \includegraphics[width=1.0\columnwidth]{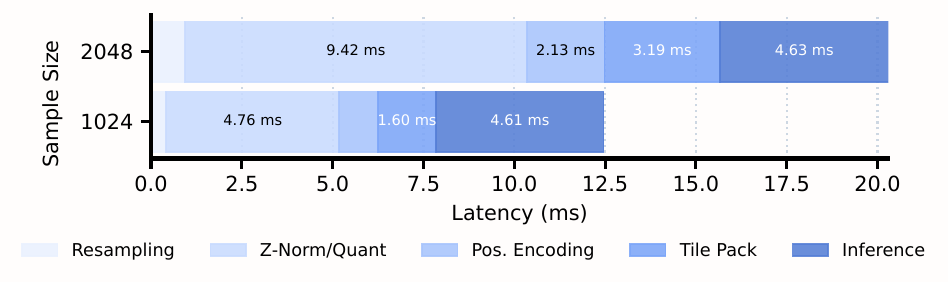}
  \vspace{-0.8cm}
  \caption{Latency breakdown on the MAX78000 for ECG input at both 1024- and 2048-sample sizes. }\label{fig:latency_breakdown}
  \vspace{-0.5cm}
\end{figure}

\begin{table}[!t]
\caption{Latency breakdown (ms) for MAX78000 and HX-WE2 across ECG and EMG inputs, including HX-WE2 CPU-only.}
\vspace{-0.3cm}
\scriptsize
\centering
\begin{tabular}{lcccccc}
\hline
\multirow{2}{*}{\textbf{Stage}} &
\multicolumn{2}{c}{\textbf{MAX78000}} &
\multicolumn{2}{c}{\textbf{HX-WE2 (NPU)}} &
\multicolumn{2}{c}{\textbf{HX-WE2 (CPU)}} \\
 & \textbf{ECG} & \textbf{EMG} & \textbf{ECG} & \textbf{EMG} & \textbf{ECG} & \textbf{EMG} \\ \hline

\multicolumn{7}{c}{\textbf{1024}} \\ \hline

Resampling     & 0.42 & 0.31 & 1.51 & 1.01 & 0.99 & 1.48 \\
Z-Norm / Quant & 4.76 & 3.11 & 3.69 & 2.45 & 2.54 & 3.82 \\
Pos. Encoding  & 1.07 & 1.15 & 0.66 & 0.66 & 0.58 & 0.58 \\
Tile / Pack    & 1.60 & 1.60 & 1.81 & 1.81 & 0.41 & 0.38 \\
Inference      & 4.61 & 4.62 & 4.04 & 4.04 & 57.8 & 57.9 \\
\textbf{End-to-End} & \textbf{12.5} & \textbf{10.8} & \textbf{11.7} & \textbf{9.97} & \textbf{62.3} & \textbf{64.1} \\ \hline

\multicolumn{7}{c}{\textbf{2048}} \\ \hline

Resampling     & 0.93 & 0.58 & 3.09 & 2.07 & 3.09 & 1.97 \\
Z-Norm / Quant & 9.42 & 6.22 & 7.35 & 4.89 & 7.60 & 5.08 \\
Pos. Encoding  & 2.13 & 2.27 & 1.31 & 1.31 & 1.15 & 1.15 \\
Tile / Pack    & 3.19 & 3.19 & 1.68 & 1.67 & 2.10 & 1.98 \\
Inference      & 4.63 & 4.63 & 4.02 & 4.02 & 57.7 & 57.8 \\
\textbf{End-to-End} & \textbf{20.3} & \textbf{16.9} & \textbf{17.5} & \textbf{13.9} & \textbf{71.6} & \textbf{68.0} \\ \hline

\end{tabular}
\label{tab:overall}
\vspace{-0.1cm}
\end{table}

\begin{table}[!t]
\caption{Power consumption (at 3.3 V in) for EMG with 1024- and 2048-sample configurations, on MAX78000 and HX-WE2. `Idle' denotes the standby state with NPU inactive and no CPU processing; `Average' is the mean power draw during inference, including CPU pre-processing; `Peak' is the maximum power draw observed during the same stage.}
\vspace{-0.2cm}
\small
\centering
\scriptsize
\begin{tabular}{llccc}
\hline
\multirow{2}{*}{\textbf{Platform}} &
\multirow{2}{*}{\textbf{State}} &
\multicolumn{2}{c}{\textbf{Power (mW)}} \\ \cline{3-4}
 & & \textbf{1024} & \textbf{2048} \\ \hline

\multirow{3}{*}{MAX78000}
 & Idle    & 13.48 & 13.44 \\
 & Average & 33.08 & 33.40 \\
 & Peak    & 34.53 & 35.65 \\ \hline

\multirow{3}{*}{HX-WE2}
 & Idle    & 88.81 & 88.60 \\
 & Average & 101.5 & 103.8 \\
 & Peak    & 111.3 & 111.0 \\ \hline

\end{tabular}
\label{tab:emg_power}
\vspace{-0.5cm}
\end{table}

\subsection{Limitations and Discussion} 
\method{} yields strong performance when compared with both compact sequence and foundation-scale models, with consistent gains from multi-scale positional encodings and distillation. Nonetheless, there remain several limitations. 

% \newpage 

First, our evaluation is limited to public benchmark datasets collected under controlled or semi-controlled conditions. Although such benchmarks are widely used, they do not capture the variability encountered in real-world deployments. In practice, physiological signals are subject to distribution shift caused by electrode-lead/sensor placement, inter-user variability, motion artifacts, signal drift, and other non-stationary noise. This motivates a user-facing evaluation under such conditions, across longitudinal, multi-session, and multi-user settings on open hardware platforms. For example, BioPulse~\cite{reddy2025biopulse} and the BioGAP family of NPU-enabled RISC-V platforms~\cite{frey2023biogap, frey2025biogapultra} provide accessible experimental platforms for embedded physiological processing. Evaluating across such platforms would also help assess portability across hardware architectures with varying operator constraints. 

Second, while our design is motivated by physiological time-series, the generality of our method remains an open question. The main components---namely, the multi-scale front-end and attention approximation---may extend to other one-dimensional inputs (e.g., inertial measurements). At present, it is not yet clear whether the measured gains stem from inductive biases specific to physiological signals (e.g., quasi-periodicity, multi-scale rhythms), or from more general representational capacity. In turn, a systematic evaluation across multiple domains would help clarify this.

\section{Conclusion}

This work introduced \method{}, a lightweight, \textmu NPU-compatible architecture for ECG/EMG modeling%that preserves the representational structure of wavelet-Transformer models while adhering to 
, designed for the strict compute, memory, and operator constraints of microcontroller-class hardware. By replacing attention with convolutional feature mixing, approximating wavelet decomposition with parallel kernels, and offloading positional encoding to the CPU, \method{} reaches near state-of-the-art on both ECG and EMG benchmarks at <10\% model size compared to recent foundation models. With sub $\sim$20~ms latency and moderate SRAM use, %and $\sim$33~mW average power draw on the MAX78000 \textmu NPU, 
real-time physiological signal analysis on battery-powered hardware appears viable.%, opening paths for interactive applications.

% \newpage

\begin{acks}
We thank our shepherd and the anonymous reviewers for their comments. This research is supported in part by Graham Taylor, an alumnus of Imperial College London, as well as by the UKRI Open Plus Fellowship (EP/W005271/1: Securing the Next Billion Consumer Devices on the Edge).
\end{acks}

%\newpage

\bibliographystyle{ACM-Reference-Format}
\bibliography{sample-base}

@ARTICLE{npu_arch,
  author={Sze, Vivienne and Chen, Yu-Hsin and Yang, Tien-Ju and Emer, Joel S.},
  journal={IEEE Solid-State Circuits Magazine}, 
  title={{How to Evaluate Deep Neural Network Processors: TOPS/W (Alone) Considered Harmful}}, 
  year={2020},
  volume={12},
  number={3},
  pages={28-41},
  doi={10.1109/MSSC.2020.3002140}}

@misc{benchmark,
      title={{Benchmarking Ultra-Low-Power $\mu$NPUs}}, 
      author={Josh Millar and Yushan Huang and Sarab Sethi and Hamed Haddadi and Anil Madhavapeddy},
      year={2025},
      eprint={2503.22567},
      archivePrefix={arXiv},
      primaryClass={cs.LG},
      url={https://arxiv.org/abs/2503.22567}, 
}

@misc{arm_ethos_u85,
  title        = {{Arm Ethos-U85 NPU}},
  author       = {{Arm Ltd.}},
  year         = {2024},
  howpublished = {\url{https://www.arm.com/products/silicon-ip-cpu/ethos/ethos-u85}},
  note         = {Accessed: 2025-11-08}
}

@manual{max78000,
  title        = {{MAX78000 Microcontroller with Ultra-Low-Power Convolutional Neural Network Accelerator}},
  author       = {{Maxim Integrated}},
  organization = {Maxim Integrated},
  year         = {2021},
  month        = may,
  note         = {Data Sheet Rev 1; 19-100868; 5/21},
  url          = {https://www.analog.com/media/en/technical-documentation/data-sheets/MAX78000.pdf}
}

@misc{mcuformer,
      title={{MCUFormer: Deploying Vision Transformers on Microcontrollers with Limited Memory}}, 
      author={Yinan Liang and Ziwei Wang and Xiuwei Xu and Yansong Tang and Jie Zhou and Jiwen Lu},
      year={2023},
      eprint={2310.16898},
      archivePrefix={arXiv},
      primaryClass={cs.CV},
      url={https://arxiv.org/abs/2310.16898}, 
}

@inproceedings{opt1,
 author = {Rusci, Manuele and Capotondi, Alessandro and Benini, Luca},
 booktitle = {Proceedings of Machine Learning and Systems},
 editor = {I. Dhillon and D. Papailiopoulos and V. Sze},
 pages = {326--335},
 title = {{Memory-Driven Mixed Low Precision Quantization for Enabling Deep Network Inference on Microcontrollers}},
 url = {https://proceedings.mlsys.org/paper_files/paper/2020/file/0c8abcf158ed12d0dd94480681186fda-Paper.pdf},
 volume = {2},
 year = {2020}
}

@inproceedings{opt2,
author = {Liberis, Edgar and Dudziak, \L{}ukasz and Lane, Nicholas D.},
title = {{$\mu$NAS: Constrained Neural Architecture Search for Microcontrollers}},
year = {2021},
isbn = {9781450382984},
publisher = {Association for Computing Machinery},
address = {New York, NY, USA},
url = {https://doi.org/10.1145/3437984.3458836},
doi = {10.1145/3437984.3458836},
booktitle = {Proceedings of the 1st Workshop on Machine Learning and Systems},
pages = {70–79},
numpages = {10},
location = {Online, United Kingdom},
series = {EuroMLSys '21}
}

@misc{opt3,
      title={{SpArSe: Sparse Architecture Search for CNNs on Resource-Constrained Microcontrollers}}, 
      author={Igor Fedorov and Ryan P. Adams and Matthew Mattina and Paul N. Whatmough},
      year={2019},
      eprint={1905.12107},
      archivePrefix={arXiv},
      primaryClass={cs.LG},
      url={https://arxiv.org/abs/1905.12107}, 
}

@misc{physiowave,
      title={{PhysioWave: A Multi-Scale Wavelet-Transformer for Physiological Signal Representation}}, 
      author={Yanlong Chen and Mattia Orlandi and Pierangelo Maria Rapa and Simone Benatti and Luca Benini and Yawei Li},
      year={2025},
      eprint={2506.10351},
      archivePrefix={arXiv},
      primaryClass={cs.LG},
      url={https://arxiv.org/abs/2506.10351}, 
}

@misc{waveformer,
      title={{WaveFormer: A Lightweight Transformer Model for sEMG-based Gesture Recognition}}, 
      author={Yanlong Chen and Mattia Orlandi and Pierangelo Maria Rapa and Simone Benatti and Luca Benini and Yawei Li},
      year={2025},
      eprint={2506.11168},
      archivePrefix={arXiv},
      primaryClass={cs.CV},
      url={https://arxiv.org/abs/2506.11168}, 
}

@article{deepecgnet,
  author    = {Manal Alghieth},
  title     = {{DeepECG-Net: a hybrid transformer-based deep learning model for real-time ECG anomaly detection}},
  journal   = {Scientific Reports},
  year      = {2025},
  volume    = {15},
  number    = {1},
  pages     = {20714},
  doi       = {10.1038/s41598-025-07781-1},
  url       = {https://doi.org/10.1038/s41598-025-07781-1},
  issn      = {2045-2322},
}

@misc{jung,
      title={{Optimizing the Deployment of Tiny Transformers on Low-Power MCUs}}, 
      author={Victor J. B. Jung and Alessio Burrello and Moritz Scherer and Francesco Conti and Luca Benini},
      year={2024},
      eprint={2404.02945},
      archivePrefix={arXiv},
      primaryClass={cs.LG},
      url={https://arxiv.org/abs/2404.02945}, 
}

@Article{dequino,
AUTHOR = {Dequino, Alberto and Bompani, Luca and Benini, Luca and Conti, Francesco},
TITLE = {{Optimizing BFloat16 Deployment of Tiny Transformers on Ultra-Low Power Extreme Edge SoCs}},
JOURNAL = {Journal of Low Power Electronics and Applications},
VOLUME = {15},
YEAR = {2025},
NUMBER = {1},
ARTICLE-NUMBER = {8},
URL = {https://www.mdpi.com/2079-9268/15/1/8},
ISSN = {2079-9268},
DOI = {10.3390/jlpea15010008}
}

@misc{adamw,
      title={{Decoupled Weight Decay Regularization}}, 
      author={Ilya Loshchilov and Frank Hutter},
      year={2019},
      eprint={1711.05101},
      archivePrefix={arXiv},
      primaryClass={cs.LG},
      url={https://arxiv.org/abs/1711.05101}, 
}

@misc{gap9,
  author       = {GreenWaves Technologies},
  title        = {{GAP9 Product Brief}},
  year         = {2021},
}

@misc{uci_emg,
  author       = {Krilova, N. and Kastalskiy, I. and Kazantsev, V. and Makarov, V. and Lobov, S.},
  title        = {EMG Data for Gestures [Dataset]},
  howpublished = {UCI Machine Learning Repository},
  year         = {2018},
  url          = {https://archive.ics.uci.edu/dataset/481/emg+data+for+gestures},
  doi          = {10.24432/C5ZP5C},
}

@misc{ptb_xl,
  author       = {Wagner, Patrick and Strodthoff, Nils and Bousseljot, Ralf-Dieter and Samek, Wojciech and Schaeffter, Tobias},
  title        = {PTB-XL, a large publicly available electrocardiography dataset (version 1.0.3)},
  howpublished = {PhysioNet},
  year         = {2022},
  url          = {https://physionet.org/content/ptb-xl/1.0.3/},
  doi          = {10.13026/kfzx-aw45},
  note         = {RRID:SCR\_007345}
}

@article{ninapro_db5,
  author       = {Pizzolato, Stefano and Tagliapietra, Luca and Cognolato, Matteo and Reggiani, Monica and Müller, Henning and Atzori, Manfredo},
  title        = {C{omparison of six electromyography acquisition setups on hand movement classification tasks}},
  journal      = {PLOS ONE},
  volume       = {12},
  number       = {10},
  pages        = {e0186132},
  year         = {2017},
  doi          = {10.1371/journal.pone.0186132},
}

@misc{epn612,
  author       = {Benalcázar, Marco E. and Barona, Lorena and Valdivieso, Leonardo and Aguas, Xavier and Zea, Jonathan},
  title        = {{EMG-EPN-612 Dataset [Dataset]}},
  howpublished = {Zenodo},
  year         = {2020},
  url          = {https://doi.org/10.5281/zenodo.4421500},
  doi          = {10.5281/zenodo.4421500},
}

@misc{chapman_shaoxing,
  author       = {Zheng, Jianwei and Guo, Hangyuan and Chu, Huimin},
  title        = {{A large scale 12-lead electrocardiogram database for arrhythmia study (version 1.0.0)}},
  howpublished = {PhysioNet},
  year         = {2022},
  url          = {https://physionet.org/content/ecg-arrhythmia/1.0.0/},
  doi          = {10.13026/wgex-er52},
  note         = {RRID:SCR\_007345}
}

@misc{cpsc_2018,
  author       = {Liu, Cheng and Ng, E.Y.-K. and others},
  title        = {{An Open Access Database for Evaluating the Algorithms of the 1st China Physiological Signal Challenge 2018}},
  howpublished = {Proceedings of the 1st China Physiological Signal Challenge (CPSC) 2018, 7th International Conference on Biomedical Engineering and Biotechnology (ICBEB 2018)},
  year         = {2018},
  url          = {https://2018.icbeb.org/Challenge.html},
  doi          = {10.1166/jmihi.2018.2442},
}

@INPROCEEDINGS{intro1,
  author={Oliver, N. and Flores-Mangas, F.},
  booktitle={International Workshop on Wearable and Implantable Body Sensor Networks (BSN'06)}, 
  title={{HealthGear: a real-time wearable system for monitoring and analyzing physiological signals}}, 
  year={2006},
  volume={},
  number={},
  pages={4 pp.-64},
  doi={10.1109/BSN.2006.27}}

@article{intro2,
  author    = {Zachi I. Attia and Suraj Kapa and Francisco Lopez-Jimenez and Paul M. McKie and Dorothy J. Ladewig and Gaurav Satam and Patricia A. Pellikka and Maurice Enriquez-Sarano and Peter A. Noseworthy and Thomas M. Munger and Samuel J. Asirvatham and Christopher G. Scott and Rickey E. Carter and Paul A. Friedman},
  title     = {{Screening for cardiac contractile dysfunction using an artificial intelligence–enabled electrocardiogram}},
  journal   = {Nature Medicine},
  year      = {2019},
  volume    = {25},
  number    = {1},
  pages     = {70--74},
  doi       = {10.1038/s41591-018-0240-2},
  url       = {https://doi.org/10.1038/s41591-018-0240-2},
  issn      = {1546-170X},
}

@article{intro3,
  author    = {Jack Tchimino and Rehne Lessmann Hansen and Peter Holmberg Jørgensen and Jakob Dideriksen and Strahinja Dosen},
  title     = {{Application of EMG feedback for hand prosthesis control in high-level amputation: a case study}},
  journal   = {Scientific Reports},
  year      = {2024},
  volume    = {14},
  number    = {1},
  pages     = {31676},
  doi       = {10.1038/s41598-024-80828-x},
  url       = {https://doi.org/10.1038/s41598-024-80828-x},
  issn      = {2045-2322},
}

@inproceedings{intro4, series={CHI ’23},
   title={{A Framework and Call to Action for the Future Development of EMG-Based Input in HCI}},
   url={http://dx.doi.org/10.1145/3544548.3580962},
   DOI={10.1145/3544548.3580962},
   booktitle={Proceedings of the 2023 CHI Conference on Human Factors in Computing Systems},
   publisher={ACM},
   author={Eddy, Ethan and Scheme, Erik J and Bateman, Scott},
   year={2023},
   month=apr, pages={1–23},
   collection={CHI ’23} }

@misc{subject_invariant,
      title={{Subject-Aware Contrastive Learning for Biosignals}}, 
      author={Joseph Y. Cheng and Hanlin Goh and Kaan Dogrusoz and Oncel Tuzel and Erdrin Azemi},
      year={2020},
      eprint={2007.04871},
      archivePrefix={arXiv},
      primaryClass={cs.LG},
      url={https://arxiv.org/abs/2007.04871}, 
}

@article{artefacts,
  author    = {Bhabesh Kalita and Nabamita Deb and Daisy Das},
  title     = {AnEEG: leveraging deep learning for effective artifact removal in EEG data},
  journal   = {Scientific Reports},
  year      = {2024},
  volume    = {14},
  number    = {1},
  pages     = {24234},
  doi       = {10.1038/s41598-024-75091-z},
  url       = {https://doi.org/10.1038/s41598-024-75091-z},
  issn      = {2045-2322},
}

@ARTICLE{ml_4_mcu_review,
  author={Saha, Swapnil Sayan and Sandha, Sandeep Singh and Srivastava, Mani},
  journal={IEEE Sensors Journal}, 
  title={{Machine Learning for Microcontroller-Class Hardware: A Review}}, 
  year={2022},
  volume={22},
  number={22},
  pages={21362-21390},
  doi={10.1109/JSEN.2022.3210773}}

@INPROCEEDINGS{myo,
  author={Benalcázar, Marco E. and Jaramillo, Andrés G. and Jonathan and Zea, A. and Páez, Andrés and Andaluz, Víctor Hugo},
  booktitle={2017 25th European Signal Processing Conference (EUSIPCO)}, 
  title={{Hand gesture recognition using machine learning and the Myo armband}}, 
  year={2017},
  volume={},
  number={},
  pages={1040-1044},
  doi={10.23919/EUSIPCO.2017.8081366}}

@misc{convnet2020s,
      title={{A ConvNet for the 2020s}}, 
      author={Zhuang Liu and Hanzi Mao and Chao-Yuan Wu and Christoph Feichtenhofer and Trevor Darrell and Saining Xie},
      year={2022},
      eprint={2201.03545},
      archivePrefix={arXiv},
      primaryClass={cs.CV},
      url={https://arxiv.org/abs/2201.03545}, 
}

@misc{patches_are_all_you_need,
      title={{Patches Are All You Need?}}, 
      author={Asher Trockman and J. Zico Kolter},
      year={2022},
      eprint={2201.09792},
      archivePrefix={arXiv},
      primaryClass={cs.CV},
      url={https://arxiv.org/abs/2201.09792}, 
}

@misc{mobilenetv2,
      title={{MobileNetV2: Inverted Residuals and Linear Bottlenecks}}, 
      author={Mark Sandler and Andrew Howard and Menglong Zhu and Andrey Zhmoginov and Liang-Chieh Chen},
      year={2019},
      eprint={1801.04381},
      archivePrefix={arXiv},
      primaryClass={cs.CV},
      url={https://arxiv.org/abs/1801.04381}, 
}

@misc{metaformer,
      title={{MetaFormer Is Actually What You Need for Vision}}, 
      author={Weihao Yu and Mi Luo and Pan Zhou and Chenyang Si and Yichen Zhou and Xinchao Wang and Jiashi Feng and Shuicheng Yan},
      year={2022},
      eprint={2111.11418},
      archivePrefix={arXiv},
      primaryClass={cs.CV},
      url={https://arxiv.org/abs/2111.11418}, 
}

@inproceedings{wavelet_conv_1,
   title={{Learning Front-end Filter-bank Parameters using Convolutional Neural Networks for Abnormal Heart Sound Detection}},
   url={http://dx.doi.org/10.1109/EMBC.2018.8512578},
   DOI={10.1109/embc.2018.8512578},
   booktitle={2018 40th Annual International Conference of the IEEE Engineering in Medicine and Biology Society (EMBC)},
   publisher={IEEE},
   author={Humayun, Ahmed Imtiaz and Ghaffarzadegan, Shabnam and Feng, Zhe and Hasan, Taufiq},
   year={2018},
   month=jul, pages={1408–1411} }

@article{wavelet_conv2,
   title={{Fully learnable deep wavelet transform for unsupervised monitoring of high-frequency time series}},
   volume={119},
   ISSN={1091-6490},
   url={http://dx.doi.org/10.1073/pnas.2106598119},
   DOI={10.1073/pnas.2106598119},
   number={8},
   journal={Proceedings of the National Academy of Sciences},
   publisher={Proceedings of the National Academy of Sciences},
   author={Michau, Gabriel and Frusque, Gaetan and Fink, Olga},
   year={2022},
   month=feb }

@inproceedings{wavelet_conv3,
  title = {Learnable Group Transform for Time-Series},
  author = {Cosentino, Romain and Aazhang, Behnaam},
  booktitle = {Proceedings of the 37th International Conference on Machine Learning},
  series = {Proceedings of Machine Learning Research},
  volume = {119},
  pages = {2164--2173},
  year = {2020},
  month = {July},
  publisher = {PMLR},
  url = {https://proceedings.mlr.press/v119/cosentino20a.html}
}

@misc{rope,
      title={{RoFormer: Enhanced Transformer with Rotary Position Embedding}}, 
      author={Jianlin Su and Yu Lu and Shengfeng Pan and Ahmed Murtadha and Bo Wen and Yunfeng Liu},
      year={2023},
      eprint={2104.09864},
      archivePrefix={arXiv},
      primaryClass={cs.CL},
      url={https://arxiv.org/abs/2104.09864}, 
}

@misc{ecgfounder,
      title={{An Electrocardiogram Foundation Model Built on over 10 Million Recordings with External Evaluation across Multiple Domains}}, 
      author={Jun Li and Aaron Aguirre and Junior Moura and Che Liu and Lanhai Zhong and Chenxi Sun and Gari Clifford and Brandon Westover and Shenda Hong},
      year={2025},
      eprint={2410.04133},
      archivePrefix={arXiv},
      primaryClass={cs.LG},
      url={https://arxiv.org/abs/2410.04133}, 
}

@misc{ecgfm,
      title={{ECG-FM: An Open Electrocardiogram Foundation Model}}, 
      author={Kaden McKeen and Sameer Masood and Augustin Toma and Barry Rubin and Bo Wang},
      year={2025},
      eprint={2408.05178},
      archivePrefix={arXiv},
      primaryClass={cs.LG},
      url={https://arxiv.org/abs/2408.05178}, 
}

@misc{himax,
  author       = {Himax Technologies},
  title        = {{WiseEye2 AI Processor}},
  year         = {2025},
  url          = {https://www.himax.com.tw/products/wiseeye-ai-sensing/wiseeye2-ai-processor/},
  note         = {Accessed: 2025-03-12}
}

@INPROCEEDINGS{physionet,
  author={Reyna, Matthew A and Sadr, Nadi and Alday, Erick A Perez and Gu, Annie and Shah, Amit J and Robichaux, Chad and Rad, Ali Bahrami and Elola, Andoni and Seyedi, Salman and Ansari, Sardar and Ghanbari, Hamid and Li, Qiao and Sharma, Ashish and Clifford, Gari D},
  booktitle={2021 Computing in Cardiology (CinC)}, 
  title={{Will Two Do? Varying Dimensions in Electrocardiography: The PhysioNet/Computing in Cardiology Challenge 2021}}, 
  year={2021},
  volume={48},
  number={},
  pages={1-4},
  doi={10.23919/CinC53138.2021.9662687}}

@article{mimic,
  author = {Gow, Brian and Pollard, Tom and Nathanson, Larry A and Johnson, Alistair and Moody, Benjamin and Fernandes, Chrystinne and Greenbaum, Nathaniel and Waks, Jonathan W and Eslami, Parastou and Carbonati, Tanner and Chaudhari, Ashish and Herbst, Elizabeth and Moukheiber, Dana and Berkowitz, Seth and Mark, Roger and Horng, Steven},
  title = {{MIMIC-IV-ECG: Diagnostic Electrocardiogram Matched Subset}},
  journal = {{PhysioNet}},
  year = {2023},
  month = sep,
  note = {Version 1.0},
  doi = {10.13026/4nqg-sb35},
  url = {https://doi.org/10.13026/4nqg-sb35}
}

@article{filtering1,
  author       = {Norbert Ádám and Dávid Val’ko and Zoltán Balogh and Branislav Madoš and Ján Hurtuk},
  title        = {{Comparative evaluation of filtration techniques for ECG signal denoising with emphasis on stationary wavelet transform}},
  journal      = {Scientific Reports},
  year         = {2025},
  volume       = {15},
  number       = {1},
  pages        = {42514},
  doi          = {10.1038/s41598-025-26476-1},
  url          = {https://doi.org/10.1038/s41598-025-26476-1},
  issn         = {2045-2322},
  publisher    = {Nature Publishing Group}
}

@article{filtering2,
title = {{Filtering the surface EMG signal: Movement artifact and baseline noise contamination}},
journal = {Journal of Biomechanics},
volume = {43},
number = {8},
pages = {1573-1579},
year = {2010},
issn = {0021-9290},
doi = {https://doi.org/10.1016/j.jbiomech.2010.01.027},
url = {https://www.sciencedirect.com/science/article/pii/S0021929010000631},
author = {Carlo J. {De Luca} and L. {Donald Gilmore} and Mikhail Kuznetsov and Serge H. Roy},
}

@misc{tflm,
      title={{TensorFlow Lite Micro: Embedded Machine Learning on TinyML Systems}}, 
      author={Robert David and Jared Duke and Advait Jain and Vijay Janapa Reddi and Nat Jeffries and Jian Li and Nick Kreeger and Ian Nappier and Meghna Natraj and Shlomi Regev and Rocky Rhodes and Tiezhen Wang and Pete Warden},
      year={2021},
      eprint={2010.08678},
      archivePrefix={arXiv},
      primaryClass={cs.LG},
      url={https://arxiv.org/abs/2010.08678}, 
}

@article {heedb,
	author = {Koscova, Zuzana and Li, Qiao and Robichaux, Chad and Moura Junior, Valdery and Ghanta, Manohar and Gupta, Aditya and Rosand, Jonathan and Aguirre, Aaron and Hong, Shenda and Albert, David E. and Xue, Joel and Parekh, Aarya and Sameni, Reza and Reyna, Matthew A. and Westover, M. Brandon and Cliford, Gari D.},
	title = {{The Harvard-Emory ECG Database}},
	elocation-id = {2024.09.27.24314503},
	year = {2024},
	doi = {10.1101/2024.09.27.24314503},
	publisher = {Cold Spring Harbor Laboratory Press},
	URL = {https://www.medrxiv.org/content/early/2024/10/01/2024.09.27.24314503},
	eprint = {https://www.medrxiv.org/content/early/2024/10/01/2024.09.27.24314503.full.pdf},
	journal = {medRxiv}
}

@article{kartsch2025riscv,
  title={{Open-source RISC-V platforms for embedded medical grade EMG processing: Are we there yet?}},
  author={Kartsch, Victor Javier and Benatti, Simone and Artoni, Fiorenzo and Micera, Silvestro and Benini, Luca},
  journal={Microprocessors and Microsystems},
  pages={105239},
  year={2025},
  publisher={Elsevier}
}

@article{frey2023biogap,
  title={{BioGAP: A 10-Core FP-Capable Ultra-Low Power IoT Processor with Medical-Grade AFE and BLE Connectivity for Wearable Biosignal Processing}},
  author={Frey, Sebastian and Guermandi, Marco and Benatti, Simone and Kartsch, Victor and Cossettini, Andrea and Benini, Luca},
  journal={arXiv preprint arXiv:2307.01619},
  year={2023}
}

@article{frey2025biogapultra,
  title={{BioGAP-Ultra: A Modular Edge-AI Platform for Wearable Multimodal Biosignal Acquisition and Processing}},
  author={Frey, Sebastian and Spacone, Giusy and Cossettini, Andrea and Guermandi, Marco and Schilk, Philipp and Benini, Luca and Kartsch, Victor},
  journal={arXiv preprint arXiv:2508.13728},
  year={2025}
}

@inproceedings{reddy2025biopulse,
  author    = {C. Rajashekar Reddy and Vivian Dsouza and Ashok Samraj Thangarajan and Przemys{\l}aw Pawe{\l}czak and Fahim Kawsar and Alessandro Montanari},
  title     = {{BioPulse: Towards Enabling Perpetual Vital Signs Monitoring Using a Body Patch}},
  booktitle = {Proceedings of the 26th International Workshop on Mobile Computing Systems and Applications (HotMobile '25)},
  pages     = {103--108},
  year      = {2025},
  publisher = {Association for Computing Machinery},
  address    = {New York, NY, USA},
  doi       = {10.1145/3708468.3711891}
}

@misc{ai8x,
  author       = {{Analog Devices Inc.}},
  title        = {ai8x-synthesis: Quantization and Synthesis for MAX78000/MAX78002 Edge AI Devices},
  year         = {2024},
  howpublished = {\url{https://github.com/analogdevicesinc/ai8x-synthesis}},
  note         = {Accessed: 2026-04-02}
}

@misc{vela,
  author       = {{Arm Ltd. and NXP Semiconductors}},
  title        = {ethos-u-vela: Neural Network Compiler for Arm Ethos-U NPUs},
  year         = {2024},
  howpublished = {\url{https://github.com/nxp-imx/ethos-u-vela}},
  note         = {Accessed: 2026-04-02}
}

@misc{monsoon,
  author = {{Monsoon Solutions Inc.}},
  title  = {High Voltage Power Monitor},
  year   = {2023},
  url    = {https://www.msoon.com/high-voltage-power-monitor},
  note   = {Accessed: 2026-04-02}
}

@inproceedings{npu_comp,
author = {Moss, Arthur and Lee, Hyunjong and Xun, Lei and Min, Chulhong and Kawsar, Fahim and Montanari, Alessandro},
title = {{Ultra-Low Power DNN Accelerators for IoT: Resource Characterization of the MAX78000}},
year = {2023},
isbn = {9781450398862},
publisher = {Association for Computing Machinery},
address = {New York, NY, USA},
url = {https://doi.org/10.1145/3560905.3568300},
doi = {10.1145/3560905.3568300},
booktitle = {Proceedings of the 20th ACM Conference on Embedded Networked Sensor Systems},
pages = {934–940},
numpages = {7},
location = {Boston, Massachusetts},
series = {SenSys '22}
}

% \newpage

% \clearpage
% \onecolumn

\appendix
%\pagenumbering{alph}

\section{Ablation Study}
\label{sec:ablations}

We conduct ablation studies to quantify the contribution of key architectural and training components, including knowledge distillation, positional encoding, and multi-scale kernel design. For single-label datasets (DB5, EPN612, UCI-EMG, PTB-XL) we report macro F\textsubscript{1}. For multilabel datasets (CPSC, Chapman-Shaoxing) we report micro F\textsubscript{1} on thresholded predictions.

\subsubsection{Impact of Distillation}
Table~\ref{tab:ablation-kd} quantifies the impact of knowledge distillation (KD). Overall, removing KD yields an average reduction of 0.038 in F\textsubscript{1} across the benchmarks (0.825$\rightarrow$0.786). The largest drops are observed on UCI (0.939$\rightarrow$0.866, $-0.073$), DB5 (0.738$\rightarrow$0.686, $-0.052$), and CPSC (0.772$\rightarrow$0.719, $-0.053$).

\subsubsection{Impact of Distillation Strength}
Table~\ref{tab:physiolite-kd-f1} quantifies the impact of distillation strength $\alpha$ on model performance. Overall, $\alpha=0.5$ achieves the best average F\textsubscript{1} across the evaluated benchmarks (0.816 vs 0.816 for $\alpha=0.3$ and 0.815 for $\alpha=0.7$). Concretely, $\alpha=0.5$ improves F\textsubscript{1} on DB5 (0.738 vs 0.734/0.735; up to $+0.003$), CPSC (0.772 vs 0.760/0.749; up to $+0.023$), and PTB-XL (0.600 vs 0.571/0.593; up to $+0.029$). Conversely, $\alpha=0.3$ performs best on EPN-612 (0.949 vs 0.939/0.938; up to $+0.011$) and Chapman-Shaoxing (0.942 vs 0.929/0.933; up to $+0.013$), while $\alpha=0.7$ achieves the highest score on UCI-EMG (0.939 vs 0.937/0.916; up to $+0.023$).

\subsubsection{Impact of Positional Encoding}
Table~\ref{tab:ablation-pe} quantifies the impact of positional encoding (PE), in the absence of KD. Using PE yield improved performance across most benchmarks, reducing \textsubscript{1} on DB5 (0.686$\rightarrow$0.638, $-0.048$), EPN-612 (0.939$\rightarrow$0.936, $-0.003$), UCI (0.866$\rightarrow$0.762, $-0.104$), PTB-XL (0.584$\rightarrow$0.566, $-0.018$), and Chapman-Shaoxing (0.923$\rightarrow$0.922, $-0.001$). Contrastingly, CPSC shows a noticeable improvement without PE (0.719$\rightarrow$0.766, $+0.047$).

\subsubsection{Impact of Multi-Scale Kernel Design}
Table~\ref{tab:ablation-kernel} quantifies the impact of our multi-scale kernel design by replacing its setup ($K={3,5,7}$) with fixed kernels of size 3, 5, and 7 (and KD $\alpha$ = 0.5). The multi-scale kernel set yields the best performance on 4/6 benchmarks. Concretely, $K={3,5,7}$ improves F\textsubscript{1} on DB5 (0.738 vs 0.706/0.704/0.593; up to $+0.145$ over $K=7$), EPN-612 (0.939 vs 0.927/0.924/0.913; up to $+0.026$), UCI (0.916 vs 0.912/0.879/0.762; up to $+0.154$), and CPSC (0.772 vs 0.755/0.746/0.755; up to $+0.026$). The exceptions are PTB-XL (0.605 with $K=3$ vs 0.600 with $K={3,5,7}$, $+0.005$) and Chapman-Shaoxing (0.938 with $K=7$ vs 0.929 with $K={3,5,7}$, $+0.009$).

% \newpage 

\begin{table}[htbp]
\centering
\caption{Impact of distillation setup on F\textsubscript{1} score across ECG and EMG benchmarks. }
\vspace{-0.3cm}
\label{tab:ablation-kd}
\resizebox{\columnwidth}{!}{
\begin{tabular}{lcccccc}
\toprule
\textbf{Model} & \textbf{DB5} & \textbf{EPN-612} & \textbf{UCI} & \textbf{PTB-XL} & \textbf{CPSC} & \textbf{Chapman-Shaoxing} \\
\midrule
\method{} & 0.738 & 0.949 & 0.939 & 0.600 & 0.772 & 0.942 \\
\method{} w/o KD & 0.686 & 0.939 & 0.866 & 0.584 & 0.719 & 0.923 \\
\bottomrule
\end{tabular}}
\vspace{-0.5cm}
\end{table}

\begin{table}[htbp]
    \centering
    \setlength{\tabcolsep}{5pt}
    \renewcommand{\arraystretch}{1.1}
    \small
    \caption{Impact of distillation strength on F\textsubscript{1} score across ECG and EMG benchmarks.}
    \vspace{-0.3cm}
    \label{tab:physiolite-kd-f1}
    \resizebox{\columnwidth}{!}{
    \begin{tabular}{l c c c c c c}
        \toprule
        \textbf{KD strength $\alpha$} & \textbf{DB5} & \textbf{EPN612} & \textbf{UCI-EMG} & \textbf{PTB-XL} & \textbf{CPSC} & \textbf{Chapman-Shaoxing} \\
        \midrule
        0.3 & 0.734 & 0.949 & 0.937 & 0.571 & 0.760 & 0.942 \\
        0.5 & 0.738 & 0.939 & 0.916 & 0.600 & 0.772 & 0.929 \\
        0.7 & 0.735 & 0.938 & 0.939 & 0.593 & 0.749 & 0.933 \\
        \bottomrule
    \end{tabular}}
\vspace{-0.5cm}    
\end{table}

\begin{table}[htbp]
\centering
\caption{Impact of positional encoding on F\textsubscript{1} score across ECG and EMG benchmarks. }
\vspace{-0.3cm}
\label{tab:ablation-pe}
\resizebox{\columnwidth}{!}{
\begin{tabular}{lcccccc}
\toprule
\textbf{Model} & \textbf{DB5} & \textbf{EPN-612} & \textbf{UCI} & \textbf{PTB-XL} & \textbf{CPSC} & \textbf{Chapman-Shaoxing} \\
\midrule
\method{} w/o KD & 0.686 & 0.939 & 0.866 & 0.584 & 0.719 & 0.923 \\
\method{} w/o KD + PE & 0.638 & 0.936 & 0.762 & 0.566 & 0.766 & 0.922 \\
\bottomrule
\end{tabular}}
\vspace{-0.5cm}
\end{table}

\begin{table}[htbp]
\centering
\caption{Impact of multi-scale kernel design on F\textsubscript{1} score across ECG and EMG benchmarks (using KD $\alpha$ = 0.5). }
\vspace{-0.3cm}
\label{tab:ablation-kernel}
\resizebox{\columnwidth}{!}{
\begin{tabular}{lcccccc}
\toprule
\textbf{Model} & \textbf{DB5} & \textbf{EPN-612} & \textbf{UCI} & \textbf{PTB-XL} & \textbf{CPSC} & \textbf{Chapman-Shaoxing} \\
\midrule
\method{} ($K=\{3,5,7\}$) & 0.738 & 0.939 & 0.916 & 0.600 & 0.772 & 0.929 \\
\method{} ($K=3$) & 0.706 & 0.927 & 0.912 & 0.605  & 0.755 &  0.932 \\
\method{} ($K=5$) & 0.704 & 0.924 & 0.879 & 0.599 & 0.746 & 0.933 \\
\method{} ($K=7$) & 0.593 & 0.913 & 0.762 & 0.593 & 0.755 & 0.938 \\
\bottomrule
\end{tabular}}
\vspace{-0.5cm}
\end{table}

% \newpage

\onecolumn

\section{Hyperparameter Configuration}
\label{sec:hyperparameters}

Table~\ref{tab:hyperparameters} summarizes our hyperparameter configuration used for all baseline models. Where applicable, hyperparameters are tuned using validation sets. Parameters not explicitly listed are left at their respective model defaults. Our setup ensures fair comparison across architectures while respecting model-specific requirements.

\begin{table*}[htbp!]
\centering
\caption{Training hyperparameter configuration for all \textit{baseline} models. Parameters not listed are left at model defaults.}
\label{tab:hyperparameters}
\scriptsize
\scalebox{0.95}{
\begin{tabular}{llllllllllll}
\toprule
\textbf{Model} &
\textbf{Filter} &
\textbf{Train Mode} &
\textbf{Batch} &
\textbf{LR} &
\textbf{Scheduler} &
\textbf{Weight} &
\textbf{Epochs} &
\textbf{Warmup} &
\textbf{Early Stop} &
\textbf{Patch Width/Size} \\
&
&
&
\textbf{Size} &
&
&
\textbf{Decay} &
\textbf{Total} &
\textbf{Epochs} &
\textbf{Patience} &
\\
\midrule
\textbf{ECG} & & & & & & & & & & \\
\midrule
ECGFounder & Manual & Finetuned (whole model) & 128 & 5e-4 & Cosine & 1e-4 & 30 & 5 & N/A & 64 \\
ECG-FM & Manual & Finetuned (linear probe) & 128 & 5e-4 & Cosine & 1e-4 & 30 & 5 & N/A & 64 \\
PhysioWave (15M) & Learned & Finetuned (whole model) & 64 & 5e-4 & Cosine & 1e-3 & 50 & 5 & 5 & 64 \\
\midrule
\textbf{EMG/sEMG} & & & & & & & & & & \\
\midrule
OTiS & Manual & Finetuned (whole model) & 128 & 5e-4 & Cosine & 1e-4 & 50 (DB5: 100) & 5 & 5 & 32 (DB5: 16) \\
WaveFormer & Manual & Scratch & 128 & 5e-4 & Cosine & 1e-4 & 50 (DB5: 100) & 0 & 3 & 32 (16 on DB5) \\
PhysioWave (5M) & Learned & Finetuned (whole model) & 64 & 5e-4 & Cosine & 1e-3 (DB5: 1e-4) & 50 (DB5: 100) & 5 & N/A & 32 \\
\midrule
\textbf{General} & & & & & & & & & & \\
\midrule
ECG/EMG-TCN & Manual & Scratch & 64 & 1e-3 & None & 1e-4 & 30 & 0 & N/A & EMG: 32, ECG: 64 (DB5: 16) \\
ECG/EGM-BiLSTM & Manual & Scratch & 64 & 1e-3 & None & 1e-4 & 30 & 0 & N/A & EMG: 32, ECG: 64 (DB5: 16) \\
\bottomrule
\end{tabular}
}
\end{table*}

\end{document}